\documentclass[11pt]{article}
\usepackage{amssymb,amsmath,graphicx}
\usepackage[utf8]{inputenc}
\usepackage[english]{babel}  
\usepackage{multirow}
\usepackage[hidelinks]{hyperref}
\usepackage{microtype}
\usepackage{enumitem}
\usepackage{afterpage,lscape}
\usepackage{booktabs}
\usepackage{xcolor}

\usepackage[norelsize,boxed,linesnumbered,noend]{algorithm2e}
\SetArgSty{textrm} 
\SetAlCapSkip{0.7em} 
\SetKwComment{Comment}{$\triangleright$\ }{}

\usepackage[round,authoryear]{natbib}
 \bibpunct[, ]{(}{)}{,}{a}{}{,}%
 \def\newblock{\ }%

\newtheorem{theorem}{Theorem}

\newcommand{\myFont}[1]{{\ttfamily\fontseries{b}\selectfont#1}}

\newcommand{\cO}{{\mathcal{O}}}
\newcommand{\cP}{{\mathcal{P}}}

\newcommand{\myblue}[1]{#1}

\usepackage{array}
\newcolumntype{H}{>{\setbox0=\hbox\bgroup}c<{\egroup}@{}}
\newcolumntype{L}[1]{>{\raggedright\let\newline\\\arraybackslash\hspace{0pt}}m{#1}}
\newcolumntype{C}[1]{>{\centering\let\newline\\\arraybackslash\hspace{0pt}}m{#1}}
\newcolumntype{R}[1]{>{\raggedleft\let\newline\\\arraybackslash\hspace{0pt}}m{#1}}

\title{Hybrid Genetic Search for the CVRP: Open-Source Implementation and SWAP* Neighborhood}

\author{Thibaut Vidal}

\begin{document}

\linespread{1.2}\selectfont

\begin{center}

\begin{LARGE}
Hybrid Genetic Search for the CVRP:\vspace*{0.3cm}\linebreak Open-Source Implementation and SWAP* Neighborhood
\end{LARGE}

\vspace*{1cm}

\textbf{Thibaut Vidal} \\
CIRRELT \& SCALE-AI Chair in Data-Driven Supply Chains \\
Department of Mathematical and Industrial Engineering, Polytechnique Montreal \\
Department of Computer Science, Pontifical Catholic University of Rio de Janeiro \\
\href{mailto:thibaut.vidal@cirrelt.ca}{\tt thibaut.vidal@cirrelt.ca}

\vspace*{0.5cm}


\vspace*{0.8cm}

\end{center}
\noindent
\textbf{Abstract.}
The vehicle routing problem is one of the most studied combinatorial optimization topics, due to its practical importance and methodological interest. Yet, despite extensive methodological progress, many recent studies are hampered by the limited access to simple and efficient open-source solution methods. Given the sophistication of current algorithms, reimplementation is becoming a difficult and time-consuming exercise that requires extensive care for details to be truly successful. Against this background, we use the opportunity of this short paper to introduce a simple ---open-source--- implementation of the hybrid genetic search (HGS) specialized to the capacitated vehicle routing problem (CVRP). This state-of-the-art algorithm uses the same general methodology as \cite{Vidal2012} but also includes additional methodological improvements and lessons learned over the past decade of research. In particular, it includes an additional neighborhood called SWAP* which consists in exchanging two customers between different routes without an insertion in place. As highlighted in our study, an efficient exploration of SWAP* moves significantly contributes to the performance of local searches. Moreover, as observed in experimental comparisons with other recent approaches on the classical instances of \cite{Uchoa2017}, HGS still stands as a leading metaheuristic regarding solution quality, convergence speed, and conceptual simplicity.
\vspace*{0.3cm}


\pagebreak

\section{Introduction}
\label{section-intro}

A decade has passed since the introduction of the hybrid genetic search with advanced diversity control (HGS in short) in \cite{Vidal2012} and the generalization of this method into a unified algorithm for the vehicle routing problem (VRP) family \citep{Vidal2012b,Vidal2014,Vidal2017b,Vidal2021}. 
Over this period, the method has produced outstanding results for an extensive collection of node and arc routing problems, and demonstrated that generality does not always hinder performance in this domain. As the VRP research community faces even more complex and integrated problems arising from e-commerce, home deliveries, and mobility-on-demand applications, efficient solution algorithms are more than ever instrumental for success~\citep{Vidal2020}. Yet, new applications and management studies are often hampered by the need for efficient and scalable routing solution methods. The repeated invention and reproduction of heuristic solution methods turns into a time-consuming exercise that typically requires extensive care for details and systematic testing to be crowned with success. Moreover, the successful application of optimization algorithms rests on a delicate equilibrium. While the methods should be as sophisticated as required, they should also remain as simple and transparent as possible.

To facilitate future studies, we use the opportunity of this short paper to introduce an open-source HGS algorithm for the canonical capacitated vehicle routing problem (CVRP). We refer to this specialized implementation as HGS-CVRP. The C++ implementation of this algorithm has been designed to be transparent, specialized, and extremely concise, retaining only the core elements that make this method a success. Indeed, we believe that without any control on code complexity and length, it is almost always possible to achieve small performance gains through additional operators and method hybridizations, but at the cost of conceptual simplicity. We strive to avoid this pitfall as intricate designs tend to hinder scientific progress, and effectively deliver an algorithm based on two complementary operators: a crossover for diversification, and efficient local search strategies for solution improvement.

Beyond a simple reimplementation of the original algorithm, HGS-CVRP takes advantage of several lessons learned from the past decade of VRP studies: it relies on simple data structures to avoid move re-evaluations and uses the optimal linear-time Split algorithm of \cite{Vidal2016}. Moreover, its specialization to the CVRP permits significant methodological simplifications. In particular, it does not rely on the visit-pattern improvement (PI) operator \citep{Vidal2012} originally designed for VRPs with multiple periods, and uses instead a new neighborhood called \textsc{Swap*}. As demonstrated in this paper, \textsc{Swap*} contains $\Theta(n^4)$ moves but can be explored in sub-quadratic time. This neighborhood contains many improving moves that otherwise would not be identified. It can also be pruned by simple geometrical arguments, and largely contributes to the search performance. Our methodological developments are backed up by detailed experimental analyses which permit to evaluate the performance of HGS-CVRP and the contribution of \textsc{Swap*}. As observed in our results, this new algorithm reaches the same solution quality as the original HGS algorithm from \citet{Vidal2012} in a fraction of its computational time, and largely outperforms all other existing CVRP algorithms.

The remainder of this paper is structured as follows.
Section~\ref{section-hgs} describes the HGS and its adaptations for the canonical CVRP. Section~\ref{section-swap} presents the new \textsc{Swap*} neighborhood. Section~\ref{section-code} quickly discusses the structure of the open-source code. Section~\ref{section-experiments} details our computational experiments, and Section~\ref{section-conclusions} finally concludes.

\section{Hybrid Genetic Search for the CVRP}
\label{section-hgs}

We consider a complete graph $G = (V,E)$ in which vertex $0$ represents a depot at which a fleet of~$m$ vehicles is based, and the remaining vertices \myblue{$\{1,\dots,n\}$} represent customer locations. Each edge $(i,j) \in E$ represents the possibility of traveling between locations $i$ and $j$ at a cost $c_{ij}$.
The CVRP consists in determining up to $m$ vehicle routes starting and ending at the depot, in such a way that each customer is visited once, the total demand of the customers in any route does not exceed the vehicle capacity $Q$, and the sum of the distances traveled by the vehicles is minimized \citep{Toth2014}.

Our modern HGS-CVRP uses the same search scheme as the original method of \cite{Vidal2012}. Its performance comes from a combination of three main strategies.
\begin{itemize}
\item A \textbf{synergistic combination of crossover-based and neighborhood-based search}, jointly evolving a population of individuals representing CVRP solutions. The former allows a diversified search in the solution space, while the latter permits aggressive solution improvement. Algorithms of this type are sometimes coined as memetic algorithms \citep{Moscato2010}.
\item A \textbf{controlled exploration of infeasible solutions}, in which any excess load in the routes is linearly penalized. This allows focusing the search in regions that are close to the feasibility boundaries, where optimal or near-optimal solutions are more likely to belong \citep{Glover2011,Vidal2013a}.
\item \textbf{Advanced population diversity management strategies} during parent and survivor selection, allowing to maintain a diversified and high-quality set of solutions and counterbalance the loss of diversity due to the neighborhood search.\\
\end{itemize}

\begin{figure}[htbp]
\vspace*{0.3cm}
\centering
 \includegraphics[width=0.88\textwidth]{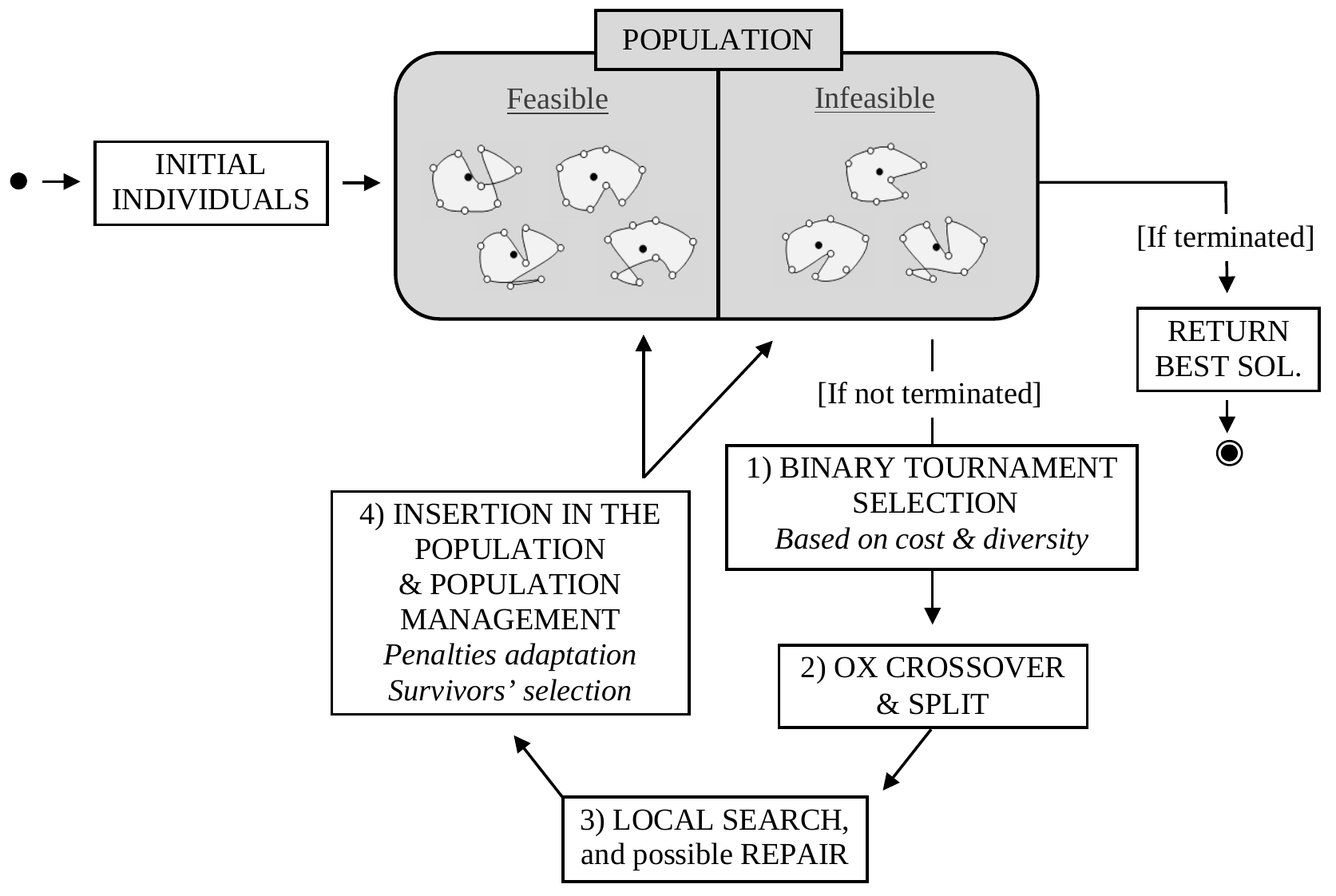}
 \caption{General structure of the hybrid genetic search}
 \label{figure-algo}
\end{figure}

The general structure of the search is represented in Figure~\ref{figure-algo}.
After a population initialization phase, the algorithm iteratively generates new solutions by 1) selecting two parents, 2) recombining them to produce a new solution, 3) improving this solution with a local search, and 4) inserting the result in the population. This process is repeated until a termination criterion is attained, typically a number of consecutive iterations $N_\textsc{it}$ without improvement or a time limit $T_\textsc{max}$.

\noindent
\textbf{Parents Selection.}
To select each parent, the algorithm performs a binary tournament selection consisting in randomly picking, with uniform probability, two individuals and retaining the one with the best fitness. It is noteworthy that the notion of fitness in HGS is based on objective value and diversity considerations. Each individual \myblue{$S$} is therefore characterized by (i) its rank \myblue{$f_\cP^\phi(S)$} in terms of solution quality, and (ii) its rank in terms of diversity contribution \myblue{$f_\cP^\textsc{div}(S)$}, measured as its average broken-pairs distance to its $n_\textsc{Closest}$ most similar solutions in the population \myblue{$\cP$}. Its fitness is then calculated as a weighted sum of these ranks \myblue{as:
\begin{equation}
f_\cP(S) = f_\cP^\phi(S) + \left(1 - \frac{n^\textsc{Elite}}{|\cP|}\right) f_\cP^\textsc{div}(S).\label{biased-fit}
\end{equation}
This equation sets a} slightly larger weight on solution quality to ensure that the top $n_\textsc{Elite}$ best individuals are preserved during the search.\\

\begin{figure}[htbp]
\centering
\includegraphics[width=0.63\textwidth]{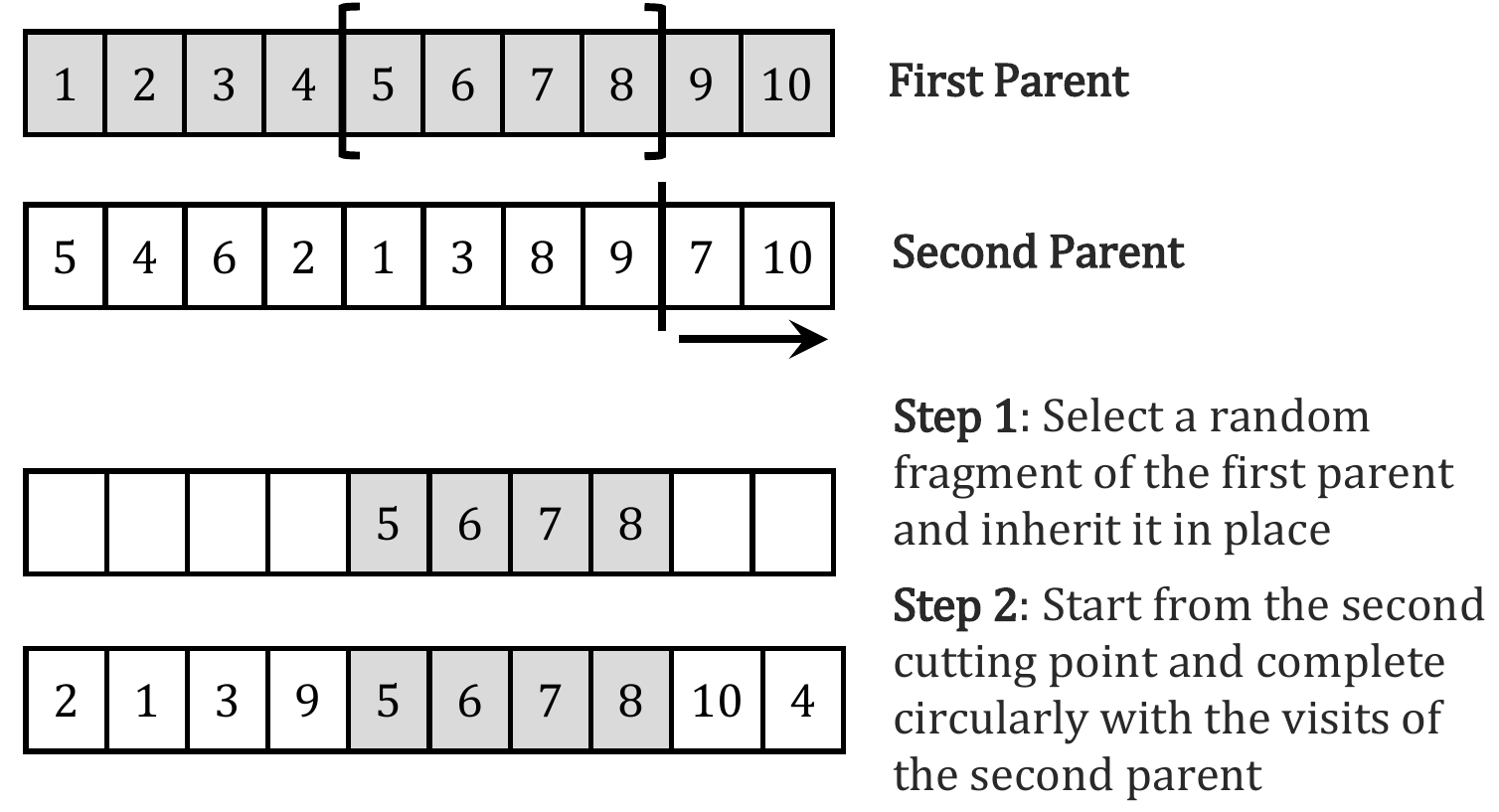}
\caption{\myblue{OX Crossover}}
\label{OX-Crossover}
\end{figure}

\noindent
\textbf{Recombination.}
HGS applies an ordered crossover (OX -- \citealt{Oliver1987}) on a simple permutation-based representation of the two parents. \myblue{As seen on Figure~\ref{OX-Crossover}, OX consists in inheriting a random fragment of the first parent, and then completing missing visits using the sequence 
of the second parent.} This representation omits the visits to the depot, in such a way that capacity constraints are disregarded in the crossover. This design choice is convenient since there exists a dynamic programming algorithm, called \textsc{Split}, capable of optimally re-inserting trip delimiters in the crossover's output to produce a complete CVRP solution \citep{Beasley1983,Prins2004}. In HGS-CVRP, we rely on the efficient linear-time \textsc{Split} algorithm introduced by \cite{Vidal2016} after each crossover operation.

\noindent
\textbf{Neighborhood Search.}
An efficient local search is applied to each solution resulting from the crossover and Split algorithms. In the original algorithm of \cite{Vidal2012}, this search included two stages: route improvement (RI) and pattern improvement (PI).
The RI local search uses \textsc{Swap} and \textsc{Relocate} moves, generalized to sequences of two consecutive nodes, as well as \textsc{2-Opt} and \textsc{2-Opt*}. The neighborhoods are limited to moves involving geographically close node pairs $(i,j)$ such that $j$ belongs to the $\Gamma$ closest clients from $i$. The \emph{granularity} parameter~$\Gamma$ therefore limits the neighborhoods' size to $\cO(\Gamma n)$. \myblue{The exploration of the moves is organized in random order of the indices $i$ and $j$ and any improving move is immediately applied. This process is pursued until attaining a local minimum.
The pattern improvement (PI) phase of \cite{Vidal2012} was originally designed to optimize the assignment of client visits to days or depots, for VRP variants with multiple periods or depots. We} excluded this mechanism in HGS-CVRP, and instead included an additional neighborhood in RI called \textsc{Swap*}, described in Section~\ref{section-swap}.

Due to the controlled exploration of infeasible solutions, is it possible for a solution to remain infeasible after the local search. When this happens, a \textsc{Repair} operation is applied with $50\%$ probability. This operation consists of running the local search with ($10\times$) higher penalty coefficients, aiming to recover a feasible solution.\\

\noindent
\textbf{Population management.}
HGS maintains two subpopulations: for feasible and infeasible solutions, respectively. Each individual produced in the previous steps is directly included in the adequate subpopulation. Each subpopulation is managed to contain between $\mu$ and $\mu + \lambda$ solutions in such a way that the parameter $\mu$ represents a minimum population size and parameter $\lambda$ is a generation size.
\myblue{Initially, $4 \mu$ random solutions are generated, improved through local search, and included in the subpopulations according to their feasibility.} Whenever a subpopulation reaches $\mu + \lambda$ individuals, a survivors selection phase is triggered to iteratively eliminate $\lambda$ solutions. This is done by iteratively removing a \emph{clone} solution (i.e., identical to another one) if such a solution exists, or otherwise the worst solution in terms of fitness \myblue{according to Equation~(\ref{biased-fit})}.

The penalty parameters for solution infeasibility are adapted through the search to achieve a target ratio $\xi^\textsc{ref}$ of feasible solutions at the end of the local search (LS). This is done by monitoring the number of feasible solutions obtained at regular intervals and increasing or decreasing the penalty coefficient by a small factor to achieve the desired target ratio, as in \cite{Vidal2012}. Finally, to deliver a method \myblue{that} remains as conceptually simple as possible, we did not include additional diversification phases in HGS-CVRP. \myblue{The complete method is summarized in~Algorithm~\ref{Algo-general}}.

\begin{algorithm}[htbp]
\myblue{
\setlength{\algomargin}{2em}
\linespread{1.05}\selectfont
 \SetAlCapSkip{0.5em}
 Initialize population with random solutions improved by local search\;
 \While{number of iterations without improvement $< It_\textsc{ni}$ and time $< T_\textsc{max}$}{
 Select parent solutions $P_1$ and $P_2$\;
 Apply the crossover operator on $P_1$ and $P_2$ to generate an offspring $C$\;
 Educate offspring $C$ by local search\;
 Insert $C$ into respective subpopulation\;
 \If{$C$ is infeasible}{
 With 50\% probability, repair $C$ (local search) and 
 insert it into respective subpopulation\;
 }
 \If{maximum subpopulation size reached}{
 Select survivors\;
 }
 Adjust penalty coefficients for infeasibility\;
 }
 Return best feasible solution\;
\caption{\textsc{HGS-CVRP}}\label{Algo-general}
}
\end{algorithm}

\section{The SWAP* Neighborhood}
\label{section-swap}

The classical \textsc{Swap} neighborhood exchanges two customers \emph{in place} (i.e., one replaces the other and vice-versa).
This neighborhood is typically used for intra-route and inter-route improvements.
In contrast, the proposed \textsc{Swap*} neighborhood consists in exchanging two customers $v$ and $v'$ from different routes $r$ and $r'$ \emph{without an insertion in place}. In this process, $v$ can be inserted in any position of $r'$, and $v'$ can likewise be inserted in any position of $r$. \myblue{Evaluating} all the \textsc{Swap*} moves would take a computational time proportional to \myblue{$\Theta(n^3)$ with a direct implementation}. However, more efficient search strategies exist. In particular, Theorem~\ref{SWAP-limitation} permits us to cut down this complexity.

\begin{theorem}
\label{SWAP-limitation}
\emph{
In a best \textsc{Swap*} move between customers $v$ and $v'$ within routes $r$ and $r'$, the new insertion position of $v$ in $r'$ is either:
\begin{enumerate}[nosep]
\item[i)] in place of $v'$, or
\item[ii)] among the three best insertion positions in $r'$ as evaluated prior to the removal of $v'$.
\end{enumerate}
A symmetrical argument holds for the new insertion position of $v'$ in $r$.
}
\end{theorem}

\textbf{Proof.} 
Let $P(r) = \{(0,r_1),(r_1,r_2),\dots,(r_{|r|-1},r_{|r|}),(r_{|r|},0)\}$ be the set of edges representing possible insertion positions in a route $r = (r_1,\dots,r_{|r|})$. The insertion cost of a vertex $v$ in a position $(i,j) \in P(r)$ is evaluated as \mbox{$\Delta(v,i,j) = c_{iv} + c_{vj} - c_{ij}$}. The best insertion cost of a vertex~$v$ in a route $r$ is calculated as
$\Delta^\textsc{min}(v,r) = \min_{(i,j) \in P(r)} \Delta(v,i,j)$.

Let $v'_\textsc{p}$ and $v'_\textsc{s}$ represent the predecessor and successor of $v'$ in its original route $r'$. After removal of $v'$, the best insertion cost of vertex $v$ in the resulting route $\hat{r}'$ is calculated as:
\begin{align}
&\Delta^\textsc{min}(v,\hat{r}') = \min \{\Delta(v,v'_\textsc{p},v'_\textsc{s}),\min_{(i,j) \in \mathcal{P}} \Delta(v,i,j)\},\label{myEq1}\\
&\text{where } \mathcal{P} = P(r')-\{(v'_\textsc{p},v'),(v',v'_\textsc{s})\}.
\end{align}
The first term of Equation~(\ref{myEq1}) represents an insertion in place of $v'$ (first statement of Theorem~\ref{SWAP-limitation}), whereas the second term corresponds to a minimum value over $P(r')$ without two elements. Therefore, this minimum is necessarily attained for one of the three best values over $P(r')$.$\hfill \Box$\\

Algorithm~\ref{algo-SWAP-Star} builds on Theorem~\ref{SWAP-limitation} to provide an efficient search strategy for \textsc{Swap*}. The neighborhood exploration is organized by route pairs $r$ and $r'$ (Lines 1--2), firstly preprocessing the three best insertion positions of each customer $v \in r$ into $r'$ (Lines 3--4) and of each customer $v \in r'$ into $r$ (Lines 5--6), and then exploiting this information to find the best move for each node $v \in r$ and $v' \in r'$ (Lines 8--16). Since the preprocessed information is valid until the routes have been modified, we use a move acceptance strategy that consists in applying the best \textsc{Swap*} move per route pair (Lines 17--18).

\begin{algorithm}[htbp]
\linespread{1.05}\selectfont
\For{each route $r \in \mathcal{R}$}
{
\For{each route $r'$ in $\Gamma^\textsc{R}(r)$}
{
\For(\Comment*[f]{Preprocessing Phase}){each customer vertex $v \in r$}
{
$((i_{1v},j_{1v}),(i_{2v},j_{2v}),(i_{3v},j_{3v})) \gets \textsc{FindTop3Locations}(v,r')$\\
}
\For{each customer vertex $v' \in r'$}
{
$((i_{1v'},j_{1v'}),(i_{2v'},j_{2v'}),(i_{3v'},j_{3v'})) \gets \textsc{FindTop3Locations}(v',r)$\\
}
$\Delta_\textsc{best} = 0$\\
\For(\Comment*[f]{Search Phase}){each customer vertex $v \in r$}
{
\For{each customer vertex $v' \in r'$}
{
$k\, = \min \{\kappa \, | \, i_{\kappa v} \neq v' \text{ and } j_{\kappa v} \neq v'\}$\\
$k' = \min \{\kappa \, | \, i_{\kappa v'} \neq v \text{ and } j_{\kappa v'} \neq v\}$\\
$\Delta_{v \rightarrow r'} =  \min \{\Delta(v,v'_\textsc{p},v'_\textsc{s}),\Delta(v,i_{kv},j_{kv})\} - \Delta(v,v_\textsc{p},v_\textsc{s})$\\
$\Delta_{v' \rightarrow r} =  \min \{\Delta(v',v_\textsc{p},v_\textsc{s}),\Delta(v',i_{k'v'},j_{k'v'})\} - \Delta(v',v'_\textsc{p},v'_\textsc{s})$\\
\If{$\Delta_{v \rightarrow r'} + \Delta_{v' \rightarrow r} < \Delta_\textsc{best}$}
{
$\Delta_\textsc{best} = \Delta_{v \rightarrow r'} + \Delta_{v' \rightarrow r}$ \\
$(v_\textsc{best},v'_\textsc{best}) = (v,v')$
}
}
}
\If{$\Delta_\textsc{best} < 0$}
{
$\textsc{ApplySwap*}(v_\textsc{best},v'_\textsc{best})$
}
}
}
\caption{Efficient exploration of the \textsc{Swap*} neighborhood}
\label{algo-SWAP-Star} 
\end{algorithm}

Each call to the function $\textsc{FindTop3Locations}(v,r)$ requires a computational time proportional to the size of the route $r$. Therefore, the computational time of Algorithm~\ref{algo-SWAP-Star} grows~as
\begin{equation}
\Phi = \sum_{r \in \mathcal{R}} \sum_{r' \in \mathcal{R}} \left( \sum_{v \in r} |r'| + \sum_{\smash{v' \in r'}} |r| + \sum_{v \in r} \sum_{\smash{v' \in r'}} 1 \right) 
= \myblue{\cO \left(\sum_{r \in \mathcal{R}} \sum_{r' \in \mathcal{R}} |r||r'| \right)}
= \myblue{\cO (n^2)}.
\end{equation}

Since a quadratic-time algorithm may still represent a bottleneck for large instances, we opted to restrict further the route pairs $(r,r')$ considered at Lines 3 and 4 using simple geometric arguments. We therefore only evaluate \textsc{Swap*} moves between $r$ and $r'$ if the polar sectors (from the depot) associated with these routes intercept each other. As shown in our computational experiments, with this additional restriction, the computational effort needed to explore \textsc{Swap*} decreases and becomes comparable to that of other standard neighborhoods in RI. \myblue{Note that the computation of polar sectors for each route requires location information for client requests (e.g., latitude/longitude or a proxy thereof). Other relatedness measures between routes (e.g., distance information or search history) could be alternatively used if such information is unavailable.}

\begin{figure}[htbp]
\centering
 \includegraphics[width=\textwidth]{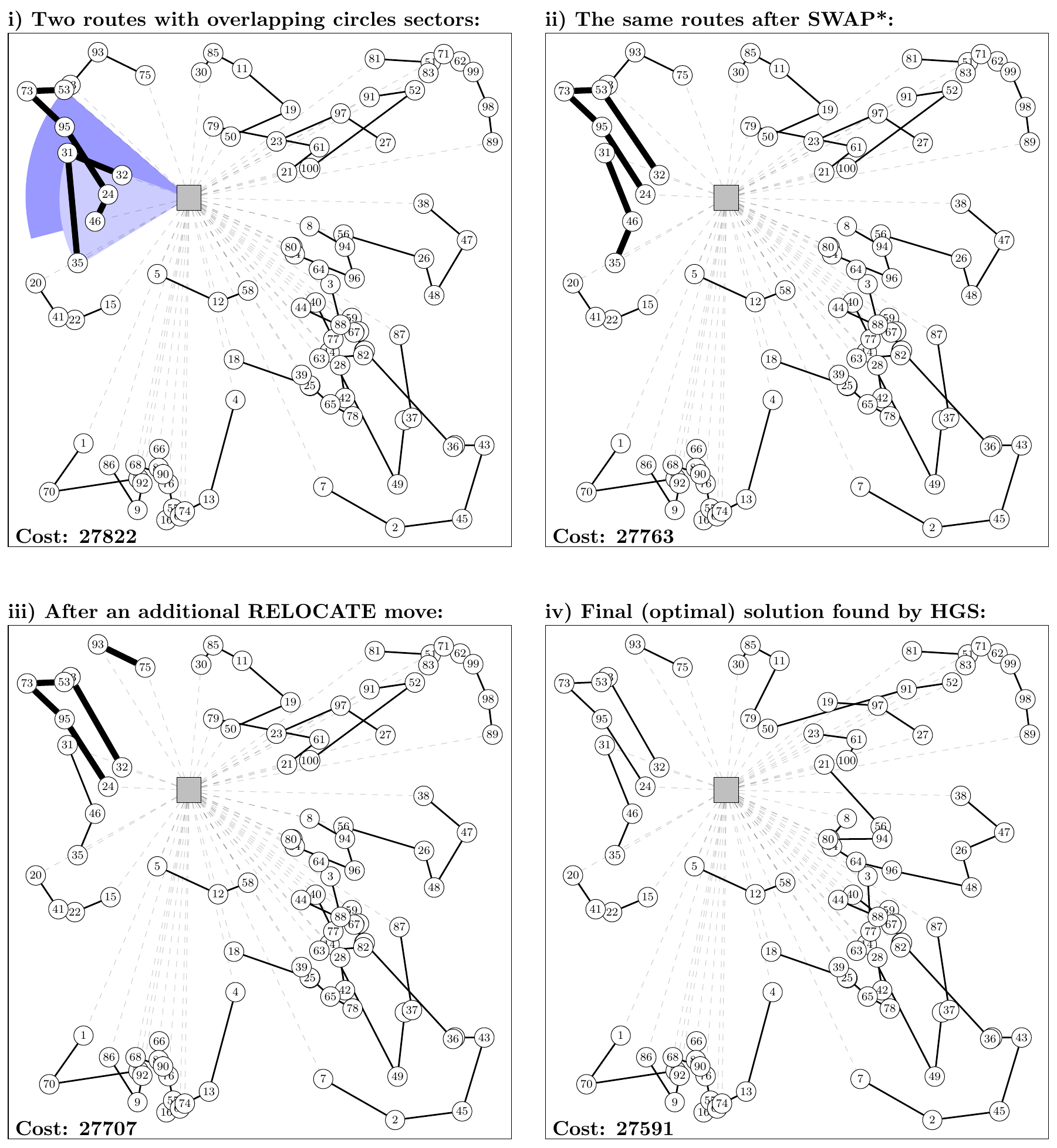}
 \caption{Illustration of the \textsc{Swap*} neighborhood on instance X-n101-k25}
 \label{figure-SWAP}
\end{figure}

Figure~\ref{figure-SWAP} illustrates the \textsc{Swap*} neighborhood during one execution of HGS-CVRP on instance X-n101-k25 from \cite{Uchoa2017}.
Four solutions are represented. The trips from and to the depot are presented with dashes to enhance readability. Solution (i) is a local minimum of all standard CVRP neighborhoods (\textsc{Swap}, \textsc{Relocate}, \textsc{2-opt} and \textsc{2-opt*}). Still, \textsc{Swap*} can identify a critical improvement between two routes highlighted in boldface on the figure. Applying this move permits to reduce the number of route intersections without violating capacity constraints. Moreover, it permits a follow-up improvement with \textsc{Relocate}, leading to Solution (iii). It is noteworthy that the solution resulting from these moves exhibits the same route arrangement in the top left quadrant as Solution (iv), which is known to be optimal for this instance.

\section{Open-Source Implementation}
\label{section-code}

Our open-source implementation is provided at
\url{https://github.com/vidalt/HGS-CVRP}. The code includes six main classes.
\begin{itemize}[nosep]
\item \myFont{[Individual]} represents the solutions (i.e., individuals of the genetic algorithm) through the search. For convenience, we store complete solutions including trip delimiters along with their associated giant tours. According to this design, the trip delimiters are immediately recalculated after crossover using the Split algorithm. This facilitates solution manipulation, feasibility checks, and distance calculations (when evaluating population diversity).
\item \myFont{[Population]} holds the two subpopulations. It also contains efficient data structures to memorize the distances between solutions used in the diversity calculations.
\item \myFont{[Genetic]} contains the main structure of the genetic algorithm and the crossover operator.
\item \myFont{[Split]} contains the linear split algorithm as introduced in \cite{Vidal2016}.
\item \myFont{[LocalSearch]} provides all the functions needed for the local search, including the \textsc{Swap*} exploration procedure which is directly embedded after the other neighborhoods. Without doubt, this is the most time-critical component of the algorithm. To efficiently represent and update the incumbent solution, we use a specialized array data structure for indexed access in $O(1)$, along with pointers giving access to the predecessors and successors in the routes. This allows efficient preprocessing, move evaluations, and solution modifications. We also use a smart data structure to register the moves that have already been tested without improvement. Instead of using binary ``move descriptors'' \citep[see, e.g.,][]{Zachariadis2010b} which require a substantial computational effort for reinitialization upon modification of a route, we rely on integer ``time stamps'' to register ``when'' a route was last modified, and ``when'' the moves associated to a given customer were last evaluated. Any non-decreasing counter can be used as a representation of time. We opted to use the number of applied moves for that purpose. A simple $O(1)$ comparison of the time stamps permits to quickly determine if the routes have been modified since the last move evaluation, and no reinitialization is needed when a route is modified.
\item \myFont{[CircleSector]} contains elementary routines to calculate polar sectors and their intersections for \textsc{Swap*}.
\item \myFont{[Params]}, \myFont{[Commandline]} and \myFont{[main]} finally store the parameters of the algorithm and permit to launch the code. The method is driven by only six parameters \myblue{summarized in Table~\ref{Parameter-Values}. All of} these parameters have been set to the original values calibrated in \cite{Vidal2012}, except $n_\textsc{Elite}$ which has been lowered down to $n_\textsc{Elite} = 4$ to favor diversity and counterbalance the additional convergence due to the \textsc{Swap*} neighborhood.

\begin{table}[htbp]
\caption{Parameters of HGS-CVRP}
\label{Parameter-Values}
\vspace*{0.1cm}
\centering
\scalebox{0.9}
{
\begin{tabular}{llr}
\toprule
\multicolumn{2}{l}{\textbf{Parameter}} & \textbf{Value}\\
\midrule
$\mu$ & Population size & 25 \\
$\lambda$ & Generation size & 40 \\
$n_\textsc{Elite}$ & Number of elite solutions considered in the fitness calculation & 4 \\
$n_\textsc{Closest}$ & Number of close solutions considered in the diversity-contribution measure & 5 \\
$\Gamma$ & Granular search parameter & 20 \\
$\xi^\textsc{ref}$ & Target proportion of feasible individuals for penalty adaptation & 0.2 \\
\bottomrule
\end{tabular}
}
\end{table}

The algorithm can be run with a termination criterion based on a number of consecutive iterations without improvement N$_\textsc{it}$ (20,000 per default) or a CPU time limit T$_\textsc{max}$. In the latter case, the algorithm restarts after each N$_\textsc{it}$ iterations without improvement and collects the best solution until the time limit.
\end{itemize}

\section{Experimental Analyses}
\label{section-experiments}

Good testing practices in optimization call for code comparisons on similar computing environments with the same number of threads (usually one) and time \citep{Talbi2009,Kendall2016a}. CPU time and solution quality cannot be examined separately, such that claims about fast solutions without a critical evaluation of solution quality are mainly inconclusive. Indeed, CPU time is often a direct consequence of the choice of termination criterion and parameters (when it is not \emph{itself} the termination criterion) and is best seen as a factor rather than an experimental outcome. 

There exist different ways to consider speed and solution quality jointly in experimental analyses. A first generic approach is to opt for a bi-objective evaluation (e.g., in Figure 4.3 of \citealt{Laporte2014a}) to identify non-dominated methods. Another intuitive approach, typical in bi-objective analyses, is to fix one dimension and measure the other, by either setting the same CPU-time limit for all algorithms or a target solution quality to achieve \citep{Aiex2007,Talbi2009}. 
Lastly, one could compare the complete convergence profiles of different algorithms during their execution by measuring solution quality over time. This approach provides the most insights, but it requires additional intervention into the algorithms to collect the solution values found throughout the search. We will favor this approach since we have access to the implementation of the algorithms being compared.

Our computational experiments follow two primary goals. 
Firstly, we evaluate the impact of our main methodological proposal ---the \textsc{Swap*} neighborhood--- by comparing two method variants: the original HGS \myblue{(referred to as HGS-2012 in the following)} with its modern HGS-CVRP implementation using \textsc{Swap*}.
Second, we extend our experimental comparison to other CVRP heuristics to compare their convergence behavior using the same computational environment.
For this analysis, we consider recent algorithms representative of the current academic state-of-the-art:
\begin{itemize}[nosep]
\item the hybrid iterated local search (HILS) of \cite{Subramanian2013};
\item \myblue{the Lin-Kernighan-Helsgaun (LKH-3) heuristic of \cite{Helsgaun2017}}; 
\item the knowledge guided local search (KGLS) of \cite{Arnold2018d};
\item the slack induction by string removals (SISR) of \cite{Christiaens2019};
\item \myblue{the fast iterative localized optimization (FILO) algorithm of \cite{Accorsi2021}}.
\end{itemize}
We were given access to the authors' implementations of HILS\myblue{, LKH-3, and FILO,} as well as to an executable library for KGLS.
The original implementation of SISR was unavailable, but we had access to a faithful reproduction of this algorithm from \cite{Reyes2020}, which achieves solutions of a quality which is statistically indistinguishable from that of the original algorithm in a similar or slightly shorter amount of time. Finally, we also establish a comparison with the solver provided by Google OR-Tools at \url{https://github.com/google/or-tools}, since this algorithm is often used in practical applications. We use the guided local search (GLS) variant of this solver as recommended in the documentation. To our knowledge, this is one of the first analyses to evaluate such a wide diversity of state-of-the-art algorithm implementations on a single test platform.

We conduct our experiments on the 100 classical benchmark instances of \cite{Uchoa2017}, as they cover diverse characteristics (e.g., distribution of demands, customer and depot positioning, route length) and represent a significant challenge for modern algorithms. HGS-2012, HGS-CVRP, HILS, \myblue{LKH-3, FILO,} and OR-Tools have been developed in \myblue{C/}C++ and compiled with g++ 9.1.0, while KGLS and SISR use Java OpenJDK 13.0.1. All experiments are run on a single thread of an Intel Gold 6148 Skylake 2.4\,GHz processor with 40\,GB of RAM, running CentOS 7.8.2003.

We monitor each algorithm's progress up to a time limit of $T_\textsc{max} = n \times 240 / 100$ seconds, where~$n$ represents the number of customers. Therefore, the smallest instance with $100$ clients is run for $4$ minutes, whereas the largest instance containing $1000$ clients is run for $40$ minutes. During each run, we record the best solution value after 1\%, 2\%, 5\%, 10\%, 15\%, 20\%, 30\%, 50\%, 75\%, and 100\% of the time limit to measure the performance of the algorithms at different stages of the search. To increase statistical significance, we perform ten independent runs with different seeds. For KGLS and OR-tools, we use different random permutations of the customers in the data file, since these two algorithms are deterministic but depend on the order of the customers in the data (in this case, the permutation of the customers effectively acts as a seed). We finally calculate the percentage gap of each algorithm as $\text{Gap} = 100 \times (z - z_\textsc{BKS})/ z_\textsc{BKS}$, where $z$ is the solution value of the algorithm and $z_\textsc{BKS}$ is the best known solution (BKS) value for this instance, as listed on the CVRPLIB website at \url{http://vrp.atd-lab.inf.puc-rio.br/index.php/en/}\footnote{Consulted on November 1$^\text{st}$, 2020}.


\begin{landscape}
\thispagestyle{empty}
\begin{table}[htbp]
\vspace*{-0.8cm}
\caption{Comparison of average solution quality over ten runs and gap at $T_\textsc{max}$}
\vspace*{0.3cm}
\setlength\tabcolsep{6.5pt}
\label{VRPres1}
\hspace*{-2cm}
\scalebox{0.73}
{
\begin{tabular}{l c rr c rr c rr c rr c rr c rr c rr c rr c r}
\toprule
\multirow{2}{*}{\textbf{Instance}}&&\multicolumn{2}{c}{\textbf{OR-Tools}}&&\multicolumn{2}{c}{\textbf{LKH-3}}&&\multicolumn{2}{c}{\textbf{HILS}}&&\multicolumn{2}{c}{\textbf{KGLS}}&&\multicolumn{2}{c}{\textbf{SISR}}&&\multicolumn{2}{c}{\textbf{FILO}}&&\multicolumn{2}{c}{\textbf{HGS-2012}}&&\multicolumn{2}{c}{\textbf{HGS-CVRP}}&&\multirow{2}{*}{\textbf{BKS}}\\
\cmidrule(lr){3-4}\cmidrule(lr){6-7}\cmidrule(lr){9-10}\cmidrule(lr){12-13}\cmidrule(lr){15-16}\cmidrule(lr){18-19}\cmidrule(lr){21-22}\cmidrule(lr){24-25}
&&\multicolumn{1}{c}{\textbf{Avg}}&\multicolumn{1}{c}{\textbf{Gap}}
&&\multicolumn{1}{c}{\textbf{Avg}}&\multicolumn{1}{c}{\textbf{Gap}}
&&\multicolumn{1}{c}{\textbf{Avg}}&\multicolumn{1}{c}{\textbf{Gap}}
&&\multicolumn{1}{c}{\textbf{Avg}}&\multicolumn{1}{c}{\textbf{Gap}}
&&\multicolumn{1}{c}{\textbf{Avg}}&\multicolumn{1}{c}{\textbf{Gap}}
&&\multicolumn{1}{c}{\textbf{Avg}}&\multicolumn{1}{c}{\textbf{Gap}}
&&\multicolumn{1}{c}{\textbf{Avg}}&\multicolumn{1}{c}{\textbf{Gap}}
&&\multicolumn{1}{c}{\textbf{Avg}}&\multicolumn{1}{c}{\textbf{Gap}}&\\
\midrule
X-n101-k25&&27977.2&1.40&&27639.2&0.17&&27591.0&0.00&&27631.9&0.15&&27593.3&0.01&&27591.0&0.00&&27591.0&0.00&&27591.0&0.00&&27591\\
X-n106-k14&&26757.5&1.50&&26406.8&0.17&&26391.1&0.11&&26413.2&0.19&&26380.9&0.07&&26373.3&0.04&&26408.8&0.18&&26381.4&0.07&&26362\\
X-n110-k13&&15099.8&0.86&&14993.9&0.15&&14971.0&0.00&&14971.0&0.00&&14972.1&0.01&&14971.0&0.00&&14971.0&0.00&&14971.0&0.00&&14971\\
X-n115-k10&&12808.3&0.48&&12747.0&0.00&&12747.0&0.00&&12747.1&0.00&&12747.0&0.00&&12747.0&0.00&&12747.0&0.00&&12747.0&0.00&&12747\\
X-n120-k6&&13501.9&1.27&&13332.8&0.01&&13333.7&0.01&&13332.0&0.00&&13332.0&0.00&&13332.0&0.00&&13332.0&0.00&&13332.0&0.00&&13332\\
X-n125-k30&&56853.4&2.37&&55907.4&0.66&&55846.5&0.55&&55740.8&0.36&&55559.8&0.04&&55693.7&0.28&&55539.0&0.00&&55539.0&0.00&&55539\\
X-n129-k18&&29722.3&2.70&&29083.3&0.50&&28972.1&0.11&&28971.6&0.11&&28948.9&0.03&&28948.4&0.03&&28940.0&0.00&&28940.0&0.00&&28940\\
X-n134-k13&&11171.0&2.34&&10970.6&0.50&&10947.5&0.29&&10940.5&0.22&&10937.7&0.20&&10927.9&0.11&&10916.0&0.00&&10916.0&0.00&&10916\\
X-n139-k10&&13741.2&1.11&&13654.9&0.48&&13591.2&0.01&&13590.0&0.00&&13590.4&0.00&&13590.0&0.00&&13590.0&0.00&&13590.0&0.00&&13590\\
X-n143-k7&&16135.6&2.77&&15767.8&0.43&&15735.7&0.23&&15730.6&0.19&&15727.8&0.18&&15723.8&0.15&&15700.0&0.00&&15700.0&0.00&&15700\\
X-n148-k46&&44598.5&2.65&&43518.9&0.16&&43448.0&0.00&&43588.3&0.32&&43464.1&0.04&&43480.5&0.07&&43448.0&0.00&&43448.0&0.00&&43448\\
X-n153-k22&&21789.3&2.68&&21240.8&0.10&&21452.3&1.09&&21386.0&0.78&&21228.6&0.04&&21232.9&0.06&&21223.5&0.02&&21225.0&0.02&&21220\\
X-n157-k13&&17137.7&1.55&&16879.1&0.02&&16876.0&0.00&&16877.5&0.01&&16878.2&0.01&&16876.0&0.00&&16876.0&0.00&&16876.0&0.00&&16876\\
X-n162-k11&&14262.2&0.88&&14173.7&0.25&&14152.4&0.10&&14147.0&0.06&&14159.0&0.15&&14157.5&0.14&&14138.0&0.00&&14138.0&0.00&&14138\\
X-n167-k10&&21176.4&3.01&&20706.1&0.73&&20603.7&0.23&&20586.9&0.15&&20558.6&0.01&&20557.0&0.00&&20557.0&0.00&&20557.0&0.00&&20557\\
X-n172-k51&&46874.9&2.78&&45788.1&0.40&&45665.3&0.13&&45802.8&0.43&&45622.6&0.03&&45607.0&0.00&&45607.0&0.00&&45607.0&0.00&&45607\\
X-n176-k26&&49260.2&3.03&&48104.1&0.61&&48218.5&0.85&&47991.6&0.38&&47823.7&0.02&&47985.0&0.36&&47812.0&0.00&&47812.0&0.00&&47812\\
X-n181-k23&&25935.6&1.43&&25627.0&0.23&&25572.1&0.01&&25602.3&0.13&&25575.1&0.02&&25569.2&0.00&&25570.2&0.00&&25569.0&0.00&&25569\\
X-n186-k15&&24908.0&3.16&&24277.7&0.55&&24170.7&0.11&&24178.3&0.14&&24166.2&0.09&&24154.6&0.04&&24145.2&0.00&&24145.0&0.00&&24145\\
X-n190-k8&&17421.9&2.60&&17074.7&0.56&&17108.0&0.75&&17033.5&0.32&&16982.8&0.02&&16984.3&0.03&&16992.4&0.07&&16983.3&0.02&&16980\\
X-n195-k51&&46151.1&4.36&&44478.8&0.57&&44305.0&0.18&&44427.2&0.46&&44292.0&0.15&&44265.7&0.09&&44225.0&0.00&&44225.0&0.00&&44225\\
X-n200-k36&&60447.9&3.19&&58913.6&0.57&&58784.0&0.35&&58828.0&0.43&&58635.6&0.10&&58806.9&0.39&&58589.6&0.02&&58578.0&0.00&&58578\\
X-n204-k19&&20348.4&4.00&&19731.3&0.85&&19617.6&0.27&&19621.0&0.29&&19653.2&0.45&&19568.4&0.02&&19565.0&0.00&&19565.0&0.00&&19565\\
X-n209-k16&&31775.5&3.65&&30925.0&0.88&&30739.0&0.27&&30709.7&0.18&&30661.7&0.02&&30684.4&0.09&&30658.7&0.01&&30656.0&0.00&&30656\\
X-n214-k11&&11374.0&4.77&&11103.4&2.28&&11077.2&2.04&&10944.3&0.81&&10894.4&0.35&&10884.3&0.26&&10877.0&0.19&&10860.5&0.04&&10856\\
X-n219-k73&&118038.0&0.38&&117669.3&0.06&&117595.0&0.00&&117689.1&0.08&&117623.7&0.02&&117595.1&0.00&&117601.7&0.01&&117596.1&0.00&&117595\\
X-n223-k34&&42046.6&3.98&&40750.9&0.78&&40549.8&0.28&&40714.4&0.69&&40535.5&0.24&&40502.8&0.16&&40455.3&0.05&&40437.0&0.00&&40437\\
X-n228-k23&&26613.4&3.39&&25879.8&0.54&&25803.7&0.24&&25836.8&0.37&&25814.3&0.28&&25781.7&0.15&&25742.7&0.00&&25742.8&0.00&&25742\\
X-n233-k16&&19883.9&3.40&&19345.8&0.60&&19296.0&0.34&&19328.6&0.51&&19285.7&0.29&&19293.9&0.33&&19233.1&0.02&&19230.0&0.00&&19230\\
X-n237-k14&&27927.5&3.27&&27164.0&0.45&&27068.8&0.10&&27095.9&0.20&&27081.1&0.14&&27050.8&0.03&&27049.4&0.03&&27042.0&0.00&&27042\\
X-n242-k48&&85518.0&3.34&&83353.0&0.73&&82867.9&0.14&&83209.2&0.55&&82885.6&0.16&&82876.1&0.15&&82826.5&0.09&&82806.0&0.07&&82751\\
X-n247-k50&&38282.8&2.71&&37412.2&0.37&&37502.3&0.61&&37388.4&0.31&&37379.6&0.28&&37453.6&0.48&&37295.0&0.06&&37277.1&0.01&&37274\\
X-n251-k28&&40087.6&3.63&&38982.0&0.77&&38859.4&0.45&&38893.3&0.54&&38765.2&0.21&&38783.5&0.26&&38735.9&0.13&&38689.9&0.02&&38684\\
X-n256-k16&&19294.5&2.42&&19086.6&1.31&&18880.8&0.22&&18891.6&0.28&&18887.3&0.26&&18880.0&0.22&&18880.0&0.22&&18839.6&0.00&&18839\\
X-n261-k13&&27920.6&5.13&&27115.6&2.10&&26808.2&0.94&&26717.5&0.60&&26595.8&0.14&&26682.4&0.47&&26594.0&0.14&&26558.2&0.00&&26558\\
\bottomrule
\end{tabular}
}
\end{table}
\end{landscape}

\begin{landscape}
\thispagestyle{empty}
\begin{table}[htbp]
\vspace*{-0.8cm}
\caption{Comparison of average solution quality over ten runs and gap at $T_\textsc{max}$ (continued)}
\vspace*{0.3cm}
\setlength\tabcolsep{6.5pt}
\label{VRPres2}
\hspace*{-2cm}
\scalebox{0.73}
{
\begin{tabular}{l c rr c rr c rr c rr c rr c rr c rr c rr c r}
\toprule
\multirow{2}{*}{\textbf{Instance}}&&\multicolumn{2}{c}{\textbf{OR-Tools}}&&\multicolumn{2}{c}{\textbf{LKH-3}}&&\multicolumn{2}{c}{\textbf{HILS}}&&\multicolumn{2}{c}{\textbf{KGLS}}&&\multicolumn{2}{c}{\textbf{SISR}}&&\multicolumn{2}{c}{\textbf{FILO}}&&\multicolumn{2}{c}{\textbf{HGS-2012}}&&\multicolumn{2}{c}{\textbf{HGS-CVRP}}&&\multirow{2}{*}{\textbf{BKS}}\\
\cmidrule(lr){3-4}\cmidrule(lr){6-7}\cmidrule(lr){9-10}\cmidrule(lr){12-13}\cmidrule(lr){15-16}\cmidrule(lr){18-19}\cmidrule(lr){21-22}\cmidrule(lr){24-25}
&&\multicolumn{1}{c}{\textbf{Avg}}&\multicolumn{1}{c}{\textbf{Gap}}
&&\multicolumn{1}{c}{\textbf{Avg}}&\multicolumn{1}{c}{\textbf{Gap}}
&&\multicolumn{1}{c}{\textbf{Avg}}&\multicolumn{1}{c}{\textbf{Gap}}
&&\multicolumn{1}{c}{\textbf{Avg}}&\multicolumn{1}{c}{\textbf{Gap}}
&&\multicolumn{1}{c}{\textbf{Avg}}&\multicolumn{1}{c}{\textbf{Gap}}
&&\multicolumn{1}{c}{\textbf{Avg}}&\multicolumn{1}{c}{\textbf{Gap}}
&&\multicolumn{1}{c}{\textbf{Avg}}&\multicolumn{1}{c}{\textbf{Gap}}
&&\multicolumn{1}{c}{\textbf{Avg}}&\multicolumn{1}{c}{\textbf{Gap}}&\\
\midrule
X-n266-k58&&77660.8&2.89&&76117.7&0.85&&75611.4&0.18&&75954.6&0.63&&75609.2&0.17&&75767.0&0.38&&75646.8&0.22&&75564.7&0.11&&75478\\
X-n270-k35&&36700.5&3.99&&35523.3&0.66&&35352.9&0.18&&35462.1&0.48&&35364.4&0.21&&35348.3&0.16&&35306.4&0.04&&35303.0&0.03&&35291\\
X-n275-k28&&22087.3&3.96&&21340.5&0.45&&21262.4&0.08&&21299.4&0.26&&21250.5&0.03&&21251.1&0.03&&21247.8&0.01&&21245.0&0.00&&21245\\
X-n280-k17&&35055.6&4.63&&33933.6&1.29&&33803.4&0.90&&33670.1&0.50&&33648.6&0.43&&33652.6&0.45&&33573.0&0.21&&33543.2&0.12&&33503\\
X-n284-k15&&21137.9&4.57&&20521.2&1.51&&20415.9&0.99&&20360.0&0.72&&20287.6&0.36&&20273.5&0.29&&20248.0&0.16&&20245.5&0.15&&20215\\
X-n289-k60&&98560.9&3.58&&96055.6&0.95&&95515.0&0.38&&95882.8&0.77&&95345.8&0.20&&95556.3&0.43&&95350.4&0.21&&95300.9&0.16&&95151\\
X-n294-k50&&49301.8&4.54&&47538.6&0.80&&47262.0&0.21&&47454.1&0.62&&47251.9&0.19&&47273.3&0.24&&47217.8&0.12&&47184.1&0.05&&47161\\
X-n298-k31&&36970.5&8.00&&34571.7&1.00&&34383.7&0.45&&34377.4&0.43&&34267.8&0.11&&34283.3&0.15&&34235.9&0.01&&34234.8&0.01&&34231\\
X-n303-k21&&22573.7&3.85&&22008.0&1.25&&21900.7&0.76&&21903.4&0.77&&21772.9&0.17&&21809.1&0.34&&21763.4&0.13&&21748.5&0.06&&21736\\
X-n308-k13&&27141.4&4.96&&26194.9&1.30&&26058.6&0.77&&26076.4&0.84&&26281.0&1.63&&25937.7&0.30&&25879.8&0.08&&25870.8&0.05&&25859\\
X-n313-k71&&97497.4&3.67&&94974.7&0.99&&94290.3&0.26&&94763.8&0.77&&94155.7&0.12&&94351.6&0.33&&94127.7&0.09&&94112.2&0.07&&94043\\
X-n317-k53&&79211.0&1.09&&78553.5&0.25&&78355.0&0.00&&78413.5&0.07&&78386.1&0.04&&78358.6&0.00&&78374.8&0.03&&78355.4&0.00&&78355\\
X-n322-k28&&31488.5&5.55&&30253.4&1.41&&29996.5&0.54&&30038.0&0.68&&29892.5&0.20&&29934.9&0.34&&29887.5&0.18&&29848.7&0.05&&29834\\
X-n327-k20&&28777.6&4.52&&27905.1&1.36&&27815.8&1.03&&27646.8&0.42&&27644.7&0.41&&27610.7&0.29&&27580.4&0.18&&27540.8&0.03&&27532\\
X-n331-k15&&32648.2&4.97&&31336.1&0.75&&31227.4&0.40&&31200.1&0.32&&31124.5&0.07&&31103.1&0.00&&31114.0&0.04&&31103.0&0.00&&31102\\
X-n336-k84&&143294.8&3.01&&140226.2&0.80&&139560.0&0.32&&140831.3&1.24&&139429.8&0.23&&139585.7&0.34&&139437.1&0.23&&139273.5&0.12&&139111\\
X-n344-k43&&44036.4&4.72&&42625.4&1.37&&42307.5&0.61&&42350.5&0.71&&42122.7&0.17&&42174.2&0.30&&42086.0&0.09&&42075.6&0.06&&42050\\
X-n351-k40&&27433.6&5.94&&26266.6&1.43&&26134.7&0.92&&26190.7&1.14&&25976.5&0.31&&25994.5&0.38&&25972.8&0.30&&25943.6&0.18&&25896\\
X-n359-k29&&53858.4&4.57&&52128.4&1.21&&52089.2&1.13&&51901.3&0.77&&51549.8&0.09&&51598.3&0.18&&51653.8&0.29&&51620.0&0.22&&51505\\
X-n367-k17&&23874.0&4.65&&23080.4&1.17&&22985.5&0.75&&22944.7&0.57&&22836.1&0.10&&22818.6&0.02&&22814.0&0.00&&22814.0&0.00&&22814\\
X-n376-k94&&148775.7&0.72&&147950.1&0.16&&147713.4&0.00&&147854.1&0.10&&147763.5&0.03&&147717.0&0.00&&147719.0&0.00&&147714.5&0.00&&147713\\
X-n384-k52&&69022.0&4.67&&66625.8&1.04&&66407.8&0.71&&66443.0&0.76&&66113.6&0.26&&66107.7&0.25&&66163.7&0.34&&66049.1&0.17&&65940\\
X-n393-k38&&40785.6&6.60&&38694.9&1.14&&38515.7&0.67&&38466.4&0.54&&38384.5&0.33&&38299.3&0.10&&38281.4&0.06&&38260.0&0.00&&38260\\
X-n401-k29&&68249.2&3.15&&66813.6&0.98&&66729.5&0.86&&66501.9&0.51&&66239.5&0.12&&66259.8&0.15&&66305.3&0.22&&66252.5&0.14&&66163\\
X-n411-k19&&20810.6&5.57&&20057.0&1.75&&19970.8&1.31&&19924.8&1.08&&19776.7&0.33&&19776.9&0.33&&19723.8&0.06&&19720.3&0.04&&19712\\
X-n420-k130&&111594.0&3.52&&108574.8&0.72&&107838.0&0.04&&108295.3&0.46&&107853.4&0.05&&107923.5&0.12&&107843.3&0.04&&107839.8&0.04&&107798\\
X-n429-k61&&68858.4&5.21&&66198.4&1.15&&65786.8&0.52&&65857.5&0.62&&65539.3&0.14&&65565.8&0.18&&65565.4&0.18&&65502.7&0.08&&65449\\
X-n439-k37&&37655.3&3.47&&36590.1&0.55&&36448.5&0.16&&36483.8&0.26&&36457.7&0.18&&36397.3&0.02&&36426.4&0.10&&36395.5&0.01&&36391\\
X-n449-k29&&58427.1&5.78&&56515.9&2.32&&56272.8&1.88&&55770.7&0.97&&55388.8&0.28&&55420.9&0.34&&55598.1&0.66&&55368.5&0.25&&55233\\
X-n459-k26&&25834.9&7.03&&24570.6&1.79&&24479.3&1.41&&24251.0&0.46&&24228.3&0.37&&24195.5&0.23&&24199.3&0.25&&24163.8&0.10&&24139\\
X-n469-k138&&230963.3&4.12&&223845.1&0.91&&222189.0&0.16&&223468.0&0.74&&222253.9&0.19&&222988.5&0.52&&222364.3&0.24&&222170.1&0.16&&221824\\
X-n480-k70&&92923.0&3.88&&90186.5&0.82&&89857.0&0.46&&89986.3&0.60&&89515.1&0.07&&89628.2&0.20&&89665.0&0.24&&89524.4&0.08&&89449\\
X-n491-k59&&70817.2&6.51&&67522.2&1.56&&67238.7&1.13&&67145.6&0.99&&66606.9&0.18&&66677.8&0.29&&66723.7&0.36&&66641.5&0.23&&66487\\
X-n502-k39&&70166.5&1.36&&69377.3&0.22&&69380.4&0.22&&69333.9&0.16&&69271.4&0.07&&69247.7&0.03&&69300.8&0.11&&69239.5&0.02&&69226\\
X-n513-k21&&25845.9&6.80&&24506.7&1.26&&24406.9&0.85&&24360.7&0.66&&24293.9&0.38&&24242.1&0.17&&24206.5&0.02&&24201.0&0.00&&24201\\
\bottomrule
\end{tabular}
}
\end{table}
\end{landscape}

\begin{landscape}
\thispagestyle{empty}
\begin{table}[htbp]
\vspace*{-0.8cm}
\caption{Comparison of average solution quality over ten runs and gap at $T_\textsc{max}$ (end)}
\vspace*{0.3cm}
\setlength\tabcolsep{6.5pt}
\label{VRPres3}
\hspace*{-2cm}
\scalebox{0.73}
{
\begin{tabular}{l c rr c rr c rr c rr c rr c rr c rr c rr c r}
\toprule
\multirow{2}{*}{\textbf{Instance}}&&\multicolumn{2}{c}{\textbf{OR-Tools}}&&\multicolumn{2}{c}{\textbf{LKH-3}}&&\multicolumn{2}{c}{\textbf{HILS}}&&\multicolumn{2}{c}{\textbf{KGLS}}&&\multicolumn{2}{c}{\textbf{SISR}}&&\multicolumn{2}{c}{\textbf{FILO}}&&\multicolumn{2}{c}{\textbf{HGS-2012}}&&\multicolumn{2}{c}{\textbf{HGS-CVRP}}&&\multirow{2}{*}{\textbf{BKS}}\\
\cmidrule(lr){3-4}\cmidrule(lr){6-7}\cmidrule(lr){9-10}\cmidrule(lr){12-13}\cmidrule(lr){15-16}\cmidrule(lr){18-19}\cmidrule(lr){21-22}\cmidrule(lr){24-25}
&&\multicolumn{1}{c}{\textbf{Avg}}&\multicolumn{1}{c}{\textbf{Gap}}
&&\multicolumn{1}{c}{\textbf{Avg}}&\multicolumn{1}{c}{\textbf{Gap}}
&&\multicolumn{1}{c}{\textbf{Avg}}&\multicolumn{1}{c}{\textbf{Gap}}
&&\multicolumn{1}{c}{\textbf{Avg}}&\multicolumn{1}{c}{\textbf{Gap}}
&&\multicolumn{1}{c}{\textbf{Avg}}&\multicolumn{1}{c}{\textbf{Gap}}
&&\multicolumn{1}{c}{\textbf{Avg}}&\multicolumn{1}{c}{\textbf{Gap}}
&&\multicolumn{1}{c}{\textbf{Avg}}&\multicolumn{1}{c}{\textbf{Gap}}
&&\multicolumn{1}{c}{\textbf{Avg}}&\multicolumn{1}{c}{\textbf{Gap}}&\\
\midrule
X-n524-k153&&156897.0&1.49&&154840.6&0.16&&155176.6&0.38&&155699.6&0.72&&154894.6&0.20&&154892.3&0.19&&154890.1&0.19&&154747.6&0.10&&154593\\
X-n536-k96&&99575.6&4.96&&96764.7&2.00&&95713.0&0.89&&95864.7&1.05&&95145.9&0.29&&95560.2&0.73&&95205.1&0.36&&95091.9&0.24&&94868\\
X-n548-k50&&89382.6&3.09&&87133.1&0.50&&86976.2&0.32&&86938.6&0.28&&86789.1&0.10&&86742.8&0.05&&86970.8&0.31&&86778.4&0.09&&86700\\
X-n561-k42&&45758.6&7.12&&43210.9&1.16&&43095.7&0.89&&43031.7&0.74&&42875.0&0.37&&42829.7&0.26&&42783.9&0.16&&42742.7&0.06&&42717\\
X-n573-k30&&52436.5&3.48&&51179.8&1.00&&51203.9&1.05&&50957.2&0.56&&50842.6&0.33&&50821.3&0.29&&50861.2&0.37&&50813.0&0.28&&50673\\
X-n586-k159&&198347.2&4.22&&191756.9&0.76&&190835.6&0.27&&191411.4&0.58&&190640.1&0.17&&190952.2&0.33&&190759.3&0.23&&190588.1&0.14&&190316\\
X-n599-k92&&113380.7&4.55&&110086.5&1.51&&109460.7&0.93&&109356.1&0.83&&108684.8&0.22&&108754.2&0.28&&108872.3&0.39&&108656.0&0.19&&108451\\
X-n613-k62&&64073.6&7.62&&60616.7&1.82&&60457.8&1.55&&60201.2&1.12&&59705.6&0.29&&59699.4&0.28&&59801.0&0.45&&59696.3&0.27&&59535\\
X-n627-k43&&64897.9&4.40&&63084.1&1.48&&63052.4&1.43&&62568.1&0.65&&62291.8&0.21&&62251.9&0.14&&62558.7&0.63&&62371.6&0.33&&62164\\
X-n641-k35&&66862.3&4.97&&64825.9&1.78&&64709.8&1.59&&64094.3&0.63&&63851.8&0.25&&63835.4&0.22&&64086.0&0.62&&63874.2&0.28&&63694\\
X-n655-k131&&107815.9&0.97&&107044.4&0.25&&106785.7&0.01&&106956.8&0.17&&106841.6&0.06&&106805.6&0.02&&106865.4&0.08&&106808.8&0.03&&106780\\
X-n670-k130&&151874.1&3.79&&147704.1&0.94&&148272.8&1.33&&147654.2&0.90&&146961.5&0.43&&147490.8&0.79&&147319.0&0.67&&146777.7&0.30&&146332\\
X-n685-k75&&74085.5&8.62&&69443.7&1.82&&68988.4&1.15&&68854.7&0.95&&68379.6&0.26&&68440.0&0.34&&68498.0&0.43&&68343.1&0.20&&68205\\
X-n701-k44&&87060.3&6.27&&83261.3&1.63&&83159.4&1.51&&82513.5&0.72&&82053.9&0.16&&82083.5&0.20&&82457.9&0.65&&82237.3&0.38&&81923\\
X-n716-k35&&46012.9&6.05&&44441.3&2.43&&44264.0&2.02&&43730.4&0.79&&43492.0&0.24&&43492.6&0.24&&43615.1&0.53&&43505.8&0.27&&43387\\
X-n733-k159&&143829.1&5.61&&137413.2&0.90&&137014.7&0.61&&137299.3&0.81&&136445.2&0.19&&136428.1&0.17&&136512.5&0.24&&136426.9&0.17&&136190\\
X-n749-k98&&82813.4&7.11&&78910.4&2.06&&78323.3&1.31&&78211.9&1.16&&77534.9&0.29&&77551.0&0.31&&77783.0&0.61&&77655.4&0.44&&77314\\
X-n766-k71&&123106.2&7.56&&116096.4&1.43&&115858.3&1.23&&115186.0&0.64&&114836.0&0.33&&114840.8&0.34&&114894.6&0.38&&114764.5&0.27&&114454\\
X-n783-k48&&77518.9&7.08&&73933.2&2.13&&73765.3&1.89&&73043.8&0.90&&72637.3&0.34&&72573.8&0.25&&73027.6&0.88&&72790.7&0.55&&72394\\
X-n801-k40&&76428.2&4.26&&74001.2&0.95&&74141.6&1.14&&73590.5&0.39&&73412.0&0.15&&73396.5&0.12&&73803.3&0.68&&73500.4&0.27&&73305\\
X-n819-k171&&165074.0&4.40&&160305.2&1.38&&159363.2&0.79&&159572.5&0.92&&158424.5&0.19&&158918.8&0.50&&158756.1&0.40&&158511.6&0.25&&158121\\
X-n837-k142&&201836.8&4.18&&195548.5&0.94&&195053.8&0.68&&195135.0&0.72&&193946.6&0.11&&194232.7&0.26&&194636.5&0.46&&194231.3&0.26&&193737\\
X-n856-k95&&91613.9&2.98&&89530.6&0.64&&89266.2&0.34&&89333.5&0.41&&89111.1&0.16&&89040.1&0.08&&89216.1&0.28&&89037.5&0.08&&88965\\
X-n876-k59&&103576.1&4.31&&100700.2&1.41&&100487.3&1.20&&100115.7&0.82&&99484.5&0.19&&99528.2&0.23&&99889.4&0.59&&99682.7&0.39&&99299\\
X-n895-k37&&58191.7&8.04&&56627.0&5.14&&55023.1&2.16&&54306.0&0.83&&54072.3&0.39&&54033.3&0.32&&54255.9&0.74&&54070.6&0.39&&53860\\
X-n916-k207&&342127.2&3.93&&331668.2&0.76&&331158.4&0.60&&331111.0&0.59&&329584.3&0.12&&330164.7&0.30&&330234.0&0.32&&329852.0&0.20&&329179\\
X-n936-k151&&140479.3&5.84&&134477.5&1.32&&135052.1&1.75&&133831.4&0.83&&133497.1&0.58&&133259.4&0.40&&133613.7&0.67&&133369.9&0.49&&132725\\
X-n957-k87&&88603.0&3.67&&86089.1&0.73&&85979.8&0.60&&85746.6&0.33&&85559.8&0.11&&85526.2&0.07&&85823.3&0.42&&85550.1&0.10&&85465\\
X-n979-k58&&123885.2&4.12&&121339.6&1.98&&120569.7&1.33&&119600.1&0.52&&119108.2&0.10&&119202.8&0.18&&119502.3&0.43&&119247.5&0.22&&118987\\
X-n1001-k43&&78084.7&7.91&&74151.1&2.48&&74158.5&2.49&&72998.9&0.88&&72533.1&0.24&&72518.9&0.22&&73051.4&0.96&&72748.8&0.54&&72359\\
\midrule
Min Gap&&&0.38&&&0.00&&&0.00&&&0.00&&&0.00&&&0.00&&&0.00&&&0.00&&\\
Avg Gap&&&4.01&&&1.00&&&0.66&&&0.53&&&0.19&&&0.20&&&0.21&&&0.11&&\\
Max Gap&&&8.62&&&5.14&&&2.49&&&1.24&&&1.63&&&0.79&&&0.96&&&0.55&&\\
\bottomrule
\end{tabular}
}
\end{table}
\end{landscape}

\begin{table}[!htbp]
\thispagestyle{empty}
\vspace*{-1.2cm}
\caption{Comparison of best solutions over ten runs at $T_\textsc{max}$}
\vspace*{0.3cm}
\setlength\tabcolsep{7pt}
\label{VRPres-Best1}
\centering
\renewcommand{\arraystretch}{0.88}
\scalebox{0.77}
{
\begin{tabular}{l C{1.5cm}C{1.5cm}C{1.5cm}C{1.5cm}C{1.5cm}C{1.5cm}C{1.5cm}C{1.5cm}C{1.5cm}}
\toprule
&\textbf{OR-Tools}&\textbf{HILS}&\textbf{LKH-3}&\textbf{KGLS}&\textbf{SISR}&\textbf{FILO}&\textbf{HGS-2012}&\textbf{HGS-CVRP}&\textbf{BKS}\\
\midrule
X-n101-k25&27865&\textbf{27591}&\textbf{27591}&27595&\textbf{27591}&\textbf{27591}&\textbf{27591}&\textbf{27591}&27591\\
X-n106-k14&26747&26381&26381&26375&26368&\textbf{26362}&26387&26364&26362\\
X-n110-k13&14986&\textbf{14971}&\textbf{14971}&\textbf{14971}&\textbf{14971}&\textbf{14971}&\textbf{14971}&\textbf{14971}&14971\\
X-n115-k10&12768&\textbf{12747}&\textbf{12747}&\textbf{12747}&\textbf{12747}&\textbf{12747}&\textbf{12747}&\textbf{12747}&12747\\
X-n120-k6&13458&\textbf{13332}&\textbf{13332}&\textbf{13332}&\textbf{13332}&\textbf{13332}&\textbf{13332}&\textbf{13332}&13332\\
X-n125-k30&56601&55701&55713&55670&\textbf{55539}&\textbf{55539}&\textbf{55539}&\textbf{55539}&55539\\
X-n129-k18&29668&28948&28954&28954&\textbf{28940}&\textbf{28940}&\textbf{28940}&\textbf{28940}&28940\\
X-n134-k13&11096&10937&10929&10930&10918&\textbf{10916}&\textbf{10916}&\textbf{10916}&10916\\
X-n139-k10&13693&13612&\textbf{13590}&\textbf{13590}&\textbf{13590}&\textbf{13590}&\textbf{13590}&\textbf{13590}&13590\\
X-n143-k7&16019&15718&15723&15726&\textbf{15700}&\textbf{15700}&\textbf{15700}&\textbf{15700}&15700\\
X-n148-k46&44334&\textbf{43448}&\textbf{43448}&43507&\textbf{43448}&\textbf{43448}&\textbf{43448}&\textbf{43448}&43448\\
X-n153-k22&21605&21225&21225&21375&21225&21225&\textbf{21220}&21225&21220\\
X-n157-k13&17086&\textbf{16876}&\textbf{16876}&\textbf{16876}&\textbf{16876}&\textbf{16876}&\textbf{16876}&\textbf{16876}&16876\\
X-n162-k11&14238&14171&\textbf{14138}&14147&\textbf{14138}&14147&\textbf{14138}&\textbf{14138}&14138\\
X-n167-k10&21158&20583&\textbf{20557}&\textbf{20557}&\textbf{20557}&\textbf{20557}&\textbf{20557}&\textbf{20557}&20557\\
X-n172-k51&46695&\textbf{45607}&\textbf{45607}&45763&\textbf{45607}&\textbf{45607}&\textbf{45607}&\textbf{45607}&45607\\
X-n176-k26&48986&47897&48140&47958&\textbf{47812}&\textbf{47812}&\textbf{47812}&\textbf{47812}&47812\\
X-n181-k23&25787&25598&\textbf{25569}&25594&\textbf{25569}&\textbf{25569}&\textbf{25569}&\textbf{25569}&25569\\
X-n186-k15&24908&24149&24147&24156&24151&24147&\textbf{24145}&\textbf{24145}&24145\\
X-n190-k8&17380&16995&17029&17001&\textbf{16980}&\textbf{16980}&16986&\textbf{16980}&16980\\
X-n195-k51&45757&44388&\textbf{44225}&44396&44241&\textbf{44225}&\textbf{44225}&\textbf{44225}&44225\\
X-n200-k36&60338&58773&58617&58756&58587&58620&\textbf{58578}&\textbf{58578}&58578\\
X-n204-k19&20212&19610&\textbf{19565}&19581&\textbf{19565}&\textbf{19565}&\textbf{19565}&\textbf{19565}&19565\\
X-n209-k16&31740&30700&30702&30685&\textbf{30656}&30659&\textbf{30656}&\textbf{30656}&30656\\
X-n214-k11&11228&11033&10917&10913&10874&10870&\textbf{10856}&\textbf{10856}&10856\\
X-n219-k73&117924&\textbf{117595}&\textbf{117595}&117651&117596&\textbf{117595}&\textbf{117595}&\textbf{117595}&117595\\
X-n223-k34&41794&40604&40490&40686&40504&40445&\textbf{40437}&\textbf{40437}&40437\\
X-n228-k23&26396&25806&25745&25808&25782&25743&\textbf{25742}&\textbf{25742}&25742\\
X-n233-k16&19682&19232&19276&19268&19232&\textbf{19230}&\textbf{19230}&\textbf{19230}&19230\\
X-n237-k14&27809&\textbf{27042}&\textbf{27042}&27044&27043&\textbf{27042}&27044&\textbf{27042}&27042\\
X-n242-k48&85518&83052&82809&83136&82805&82775&82771&82771&82751\\
X-n247-k50&37853&37292&37300&37317&\textbf{37274}&\textbf{37274}&37278&\textbf{37274}&37274\\
X-n251-k28&40007&38918&38798&38847&38687&38723&38699&\textbf{38684}&38684\\
X-n256-k16&19067&18986&18880&18888&18880&18880&18880&\textbf{18839}&18839\\
X-n261-k13&27760&26844&26692&26671&\textbf{26558}&26612&26570&\textbf{26558}&26558\\
X-n266-k58&77275&75855&\textbf{75478}&75793&75549&75664&75608&\textbf{75478}&75478\\
X-n270-k35&36401&35432&35324&35447&35325&35309&35303&35303&35291\\
X-n275-k28&21918&21257&\textbf{21245}&21265&\textbf{21245}&\textbf{21245}&\textbf{21245}&\textbf{21245}&21245\\
X-n280-k17&34859&33690&33725&33598&33545&33608&33542&33506&33503\\
X-n284-k15&20872&20373&20325&20323&20261&20257&20225&20231&20215\\
X-n289-k60&97868&95754&95401&95770&95245&95429&95181&95242&95151\\
X-n294-k50&49010&47430&47240&47413&47199&47240&47210&47167&47161\\
X-n298-k31&36296&34391&34318&34359&34234&34234&\textbf{34231}&\textbf{34231}&34231\\
X-n303-k21&22376&21878&21806&21845&21753&21792&21754&21739&21736\\
X-n308-k13&26934&25992&25989&25999&26224&25862&25865&25862&25859\\
X-n313-k71&96958&94778&94216&94652&94098&94246&94050&94045&94043\\
X-n317-k53&78863&78408&\textbf{78355}&78391&78361&\textbf{78355}&\textbf{78355}&\textbf{78355}&78355\\
X-n322-k28&30932&30078&29923&30010&29861&29878&29857&\textbf{29834}&29834\\
X-n327-k20&28592&27786&27767&27613&27611&27565&27576&\textbf{27532}&27532\\
X-n331-k15&32493&31153&31136&31111&31122&31103&31107&\textbf{31102}&31102\\
\bottomrule
\end{tabular}
}
\end{table}

\begin{table}[!htbp]
\thispagestyle{empty}
\vspace*{-1.2cm}
\caption{Comparison of best solutions over ten runs at $T_\textsc{max}$ (end)}
\vspace*{0.3cm}
\setlength\tabcolsep{7pt}
\label{VRPres-Best2}
\centering
\renewcommand{\arraystretch}{0.88}
\scalebox{0.77}
{
\begin{tabular}{l C{1.5cm}C{1.5cm}C{1.5cm}C{1.5cm}C{1.5cm}C{1.5cm}C{1.5cm}C{1.5cm}C{1.5cm}}
\toprule
&\textbf{OR-Tools}&\textbf{HILS}&\textbf{LKH-3}&\textbf{KGLS}&\textbf{SISR}&\textbf{FILO}&\textbf{HGS-2012}&\textbf{HGS-CVRP}&\textbf{BKS}\\
\midrule
X-n336-k84&142905&139655&139351&140716&139272&139324&139305&139205&139111\\
X-n344-k43&43560&42450&42190&42229&42081&42089&42067&42061&42050\\
X-n351-k40&27093&26142&26050&26150&25965&25960&25948&25924&25896\\
X-n359-k29&53541&51852&51820&51662&51514&51514&51598&51566&51505\\
X-n367-k17&23597&22959&22956&22867&22821&\textbf{22814}&\textbf{22814}&\textbf{22814}&22814\\
X-n376-k94&148630&147876&\textbf{147713}&147801&147736&\textbf{147713}&\textbf{147713}&\textbf{147713}&147713\\
X-n384-k52&68550&66489&66351&66363&66046&66036&66088&65997&65940\\
X-n393-k38&40303&38607&38360&38433&38338&38290&\textbf{38260}&\textbf{38260}&38260\\
X-n401-k29&67913&66584&66597&66466&66222&66227&66256&66209&66163\\
X-n411-k19&20571&19860&19834&19782&19757&19758&19718&19716&19712\\
X-n420-k130&110857&108292&107801&108175&107809&107826&107831&107810&107798\\
X-n429-k61&68113&65939&65689&65795&65494&65509&65524&65484&65449\\
X-n439-k37&37171&36491&36402&36445&36402&36395&36413&36395&36391\\
X-n449-k29&58066&56212&56154&55675&55296&55358&55484&55306&55233\\
X-n459-k26&25435&24479&24363&24234&24187&24157&24182&\textbf{24139}&24139\\
X-n469-k138&230460&223289&221939&223086&222090&222543&222131&221916&221824\\
X-n480-k70&92457&90034&89629&89926&89458&89540&89561&89498&89449\\
X-n491-k59&69944&67280&67098&67034&66502&66605&66653&66569&66487\\
X-n502-k39&70032&69275&69340&69307&69238&69227&69264&69230&69226\\
X-n513-k21&25295&24384&24316&24293&24237&\textbf{24201}&\textbf{24201}&\textbf{24201}&24201\\
X-n524-k153&156322&154657&154656&155422&154758&154610&154713&154646&154593\\
X-n536-k96&98815&96214&95663&95781&95071&95485&95125&95040&94868\\
X-n548-k50&89066&87059&86813&86901&86710&86707&86888&86710&86700\\
X-n561-k42&45330&43070&42918&42989&42799&42756&42738&42726&42717\\
X-n573-k30&52080&51013&51102&50849&50777&50757&50797&50757&50673\\
X-n586-k159&197853&191412&190695&191260&190454&190865&190576&190470&190316\\
X-n599-k92&112831&109646&109119&109125&108598&108654&108712&108605&108451\\
X-n613-k62&63561&60217&60333&60111&59609&59584&59698&59636&59535\\
X-n627-k43&64590&62755&62928&62486&62221&62228&62377&62238&62164\\
X-n641-k35&66652&64638&64591&63952&63802&63769&63976&63782&63694\\
X-n655-k131&107710&106970&\textbf{106780}&106936&106808&\textbf{106780}&106841&106785&106780\\
X-n670-k130&151071&147139&147974&147477&146676&147247&146833&146640&146332\\
X-n685-k75&73090&69234&68843&68628&68271&68355&68414&68288&68205\\
X-n701-k44&86604&82863&82982&82447&82007&82006&82302&82075&81923\\
X-n716-k35&45704&44074&44058&43627&43449&43461&43541&43459&43387\\
X-n733-k159&142650&137172&136848&137185&136344&136317&136405&136323&136190\\
X-n749-k98&82083&78612&78213&78109&77399&77467&77671&77563&77314\\
X-n766-k71&121645&115732&115396&115011&114751&114703&114834&114679&114454\\
X-n783-k48&76764&73718&73512&72974&72544&72486&72797&72704&72394\\
X-n801-k40&76262&73849&73970&73500&73362&73322&73678&73396&73305\\
X-n819-k171&164377&159697&159261&159396&158344&158661&158639&158391&158121\\
X-n837-k142&201518&195308&194857&194988&193868&194142&194444&194103&193737\\
X-n856-k95&91109&89327&89082&89218&89042&88996&89110&88986&88965\\
X-n876-k59&103017&100539&100418&100048&99405&99421&99765&99596&99299\\
X-n895-k37&57607&56334&54856&54240&53982&53966&54134&54023&53860\\
X-n916-k207&340947&331018&330920&331006&329418&329882&329918&329572&329179\\
X-n936-k151&139456&133944&134732&133713&133190&132999&133102&133121&132725\\
X-n957-k87&88222&85893&85864&85656&85493&85493&85709&85506&85465\\
X-n979-k58&123379&120791&120142&119559&119065&119145&119353&119180&118987\\
X-n1001-k43&77117&73994&73749&72882&72414&72443&72824&72678&72359\\
\midrule
Nb BKS&0&9&20&6&21&29&34&44&\\
\bottomrule
\end{tabular}
}
\end{table}

\clearpage

\noindent
\textbf{Final solution quality.}
Tables~\ref{VRPres1}--\ref{VRPres3} compare the solution quality of the different algorithms at~$T_\textsc{max}$. For each method and instance, these tables indicate the average solution value found over the ten runs, as well as the average percentage gap from the BKS. The bottom lines of the last table also report a summary of the gaps over all instances. \myblue{Moreover, Tables~\ref{VRPres-Best1}--\ref{VRPres-Best2} summarize the best solutions found by the methods over the ten runs. Solutions matching the BKS are highlighted in boldface.}

As visible in these experiments, HGS-CVRP obtains (with an average gap of $0.11\%$ at the completion of the runs) solutions of significantly higher quality than the other approaches, followed by SISR ($0.19\%$), \myblue{FILO ($0.20\%$),} HGS-2012 ($0.21\%$), KGLS ($0.53\%$), HILS ($0.66\%$), \myblue{LKH-3 ($1.00\%$),} and OR-Tools ($4.01\%$). The original HGS-2012 found good quality solutions, but the inclusion of the \textsc{Swap*} neighborhood has led to a significant methodological breakthrough, roughly reducing the remaining gap by half. The small and medium instances especially (first 50 instances with up to 330 customers) are always solved to optimality or near-optimality, with an average gap of 0.02\% for HGS-CVRP.
Despite their unquestionable ease of configuration, flexibility, and usefulness for practical settings, \myblue{LKH-3 and} OR-Tools \myblue{are} generally outperformed by the other state-of-the-art algorithms, \myblue{with 1\% and} 4\% excess distance \myblue{respectively}. These are significant differences given the tight profit margins of the transportation sector. In terms of the number of BKS, HGS-CVRP leads with $44$ BKS, followed by HGS-2012 ($34$), \myblue{FILO ($29$),} SISR ($21$), HILS (18), \myblue{LKH-3 (9),} KGLS (6), and finally OR-Tools (0). Gathering the best solutions over all our experiments has led to 17 new BKS, which have been added to the CVRPlib on September 21$^\text{th}$, 2020.\\

\begin{figure}[htbp]
\centering
\includegraphics[width=0.72\textwidth]{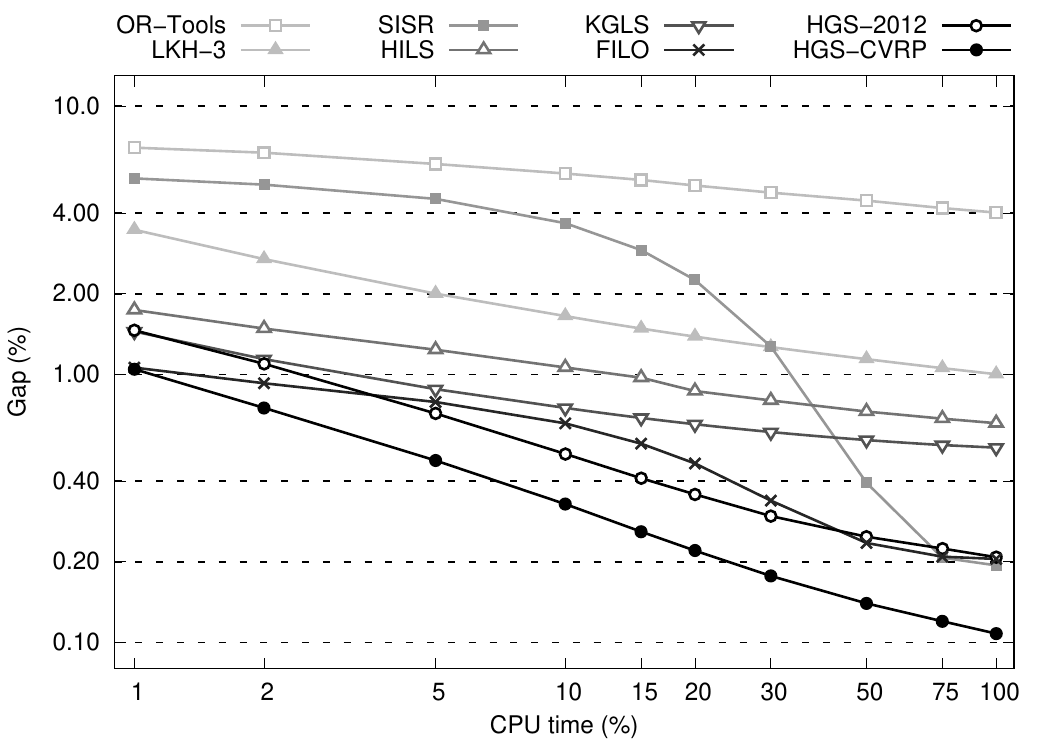}\vspace*{-0.2cm}
\caption{Convergence of the algorithms over time}
\label{Convergence-General}
\end{figure}

\begin{figure}[htbp]
\hspace*{-1cm}
\includegraphics[width=1.1\textwidth]{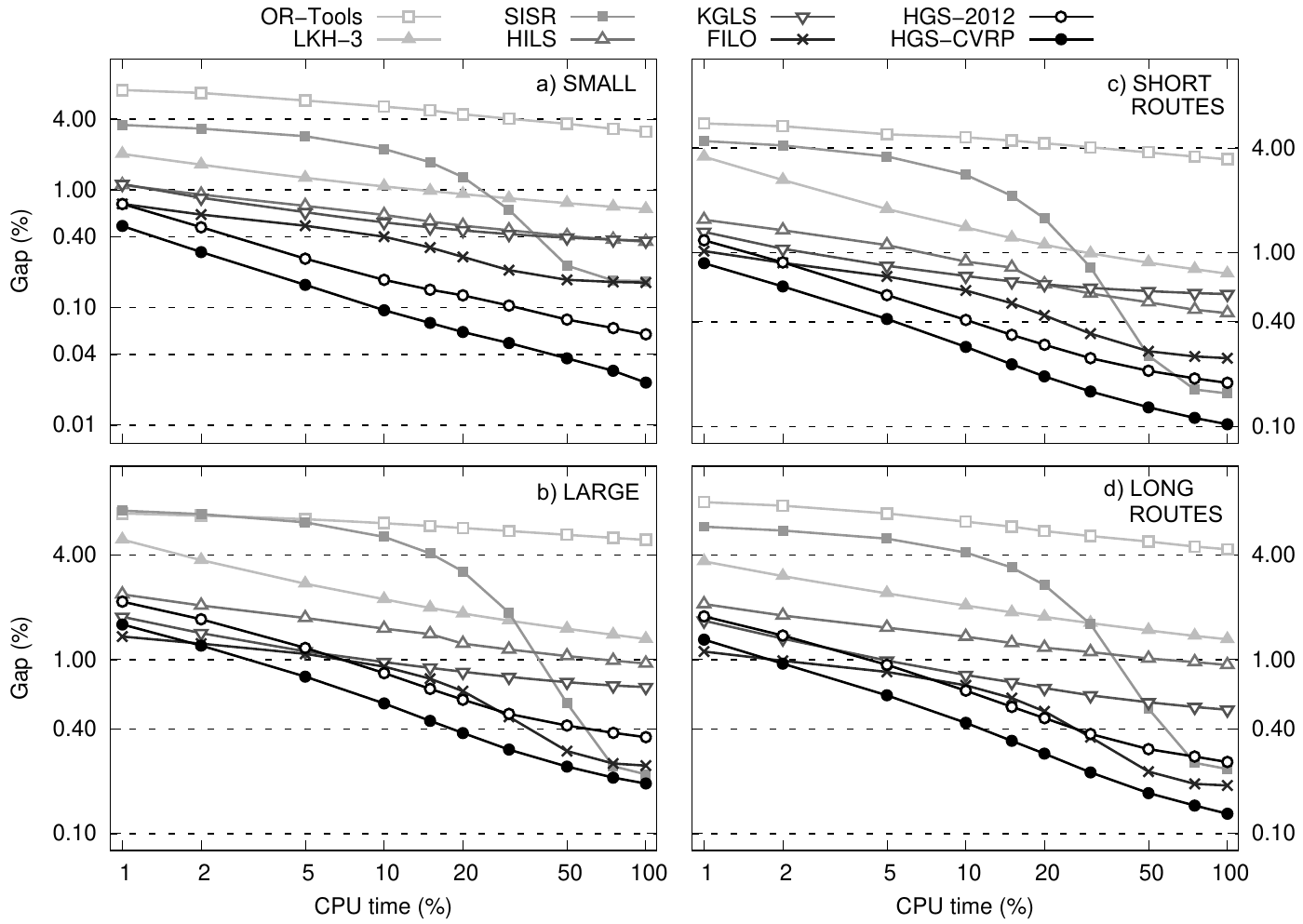}
\caption{Convergence of the algorithms over time for different subgroups of instances}
\label{Convergence-Multiple}
\end{figure}

\noindent
\textbf{Convergence over time.}
Figures~\ref{Convergence-General}~and~\ref{Convergence-Multiple} show the algorithms' progress over time on a logarithmic scale. The former figure is based on the results of all instances, whereas the latter figure focuses on subsets of these instances:
\begin{itemize}[nosep]
\item a) \textsc{Small}: First 50 instances counting between 100 and 330 customers;
\item b) \textsc{Large}: Last 50 instances counting between 335 and 1000 customers;
\item c) \textsc{Short Routes}: Instances of index $i = 5k+1$ or $i = 5k+2$ for $k \in \mathbb{N}$. By design, these~$40$ instances have a smaller number of customers per route \citep{Uchoa2017}.
\item d) \textsc{Long Routes}:  Instances of index $i = 5k+4$ or $i = 5k$ for $k \in \mathbb{N}$. By design, these~$40$ instances have a larger number of customers per route \citep{Uchoa2017}.
\end{itemize}

As visible in Figures~\ref{Convergence-General}--\ref{Convergence-Multiple}, the relative rank of the algorithms in terms of solution quality remains generally stable over time, except for SISR which converges towards high-quality solutions only later in the run. This behavior is due to its simulated-annealing acceptance criterion (with an exponential temperature decay), which favors exploration during the earlier phases of the search and only triggers convergence when the temperature is low enough. We observe that HGS-CVRP outperforms HGS-2012 and the other algorithms at any point in time (from 1\% to 100\%) and that \textsc{Swap*} positively impacts the search even at early stages. KGLS finds good quality solutions in general, but it does not converge faster than other approaches as initially claimed. HILS performs better on instances with short routes, due to the increased effectiveness of its set-partitioning component in that regime. Finally, SISR \myblue{and FILO} achieve their peak performance on larger instances, confirming the observations of the authors.\\

\begin{figure}[htbp]
\hspace*{-1cm}
\includegraphics[width=1.12\textwidth]{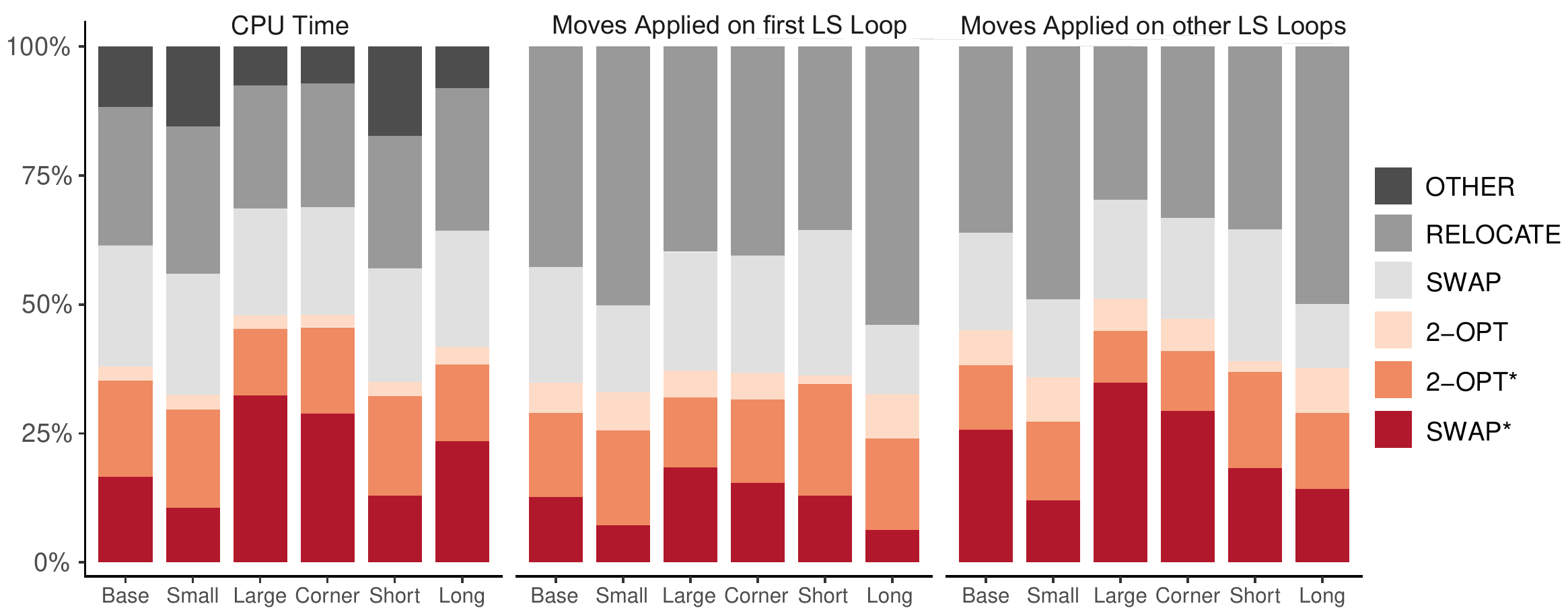}\vspace*{-0.4cm}
\caption{Statistics on the use of the neighborhoods of HGS-CVRP}
\label{Distribution-Time}
\end{figure}

\noindent
\textbf{Analysis of the SWAP* neighborhood.}
To conclude our analysis, Figure~\ref{Distribution-Time} displays the share of the computational time dedicated to the exploration of each neighborhood (\textsc{Relocate} and variants thereof involving consecutive node pairs, \textsc{Swap} and variants thereof involving consecutive node pairs,  \textsc{2-Opt}, \textsc{2-Opt*}, and \textsc{Swap*}), as well as the relative proportion of moves of each type applied during the first loop of the local search (the first time these moves are tried for a given solution) and during the rest of the local search (once the most ``obvious'' local search moves have been already applied). These figures are based on run statistics collected during ten executions on six instances with diverse characteristics from the extended ``sensitivity analysis'' benchmark of \cite{Uchoa2017}. The characteristics of these instances are summarized in Table~\ref{Sensitiv-Characteristics}, in which~$r$ represents the expected number of customers in the routes.

\begin{table}[htbp]
\caption{Characteristics of the instances used in the sensitivity analysis}
\label{Sensitiv-Characteristics}
\vspace*{0.1cm}
\centering
\scalebox{0.9}
{
\begin{tabular}{lccc}
\toprule
Instance&$n$&Depot&$r$\\
\midrule
Base&200&Centered& [9,11]\\
Small&100&Centered& [9,11]\\
Large&600&Centered& [9,11]\\
Corner&200&Corner& [9,11]\\
Short&200&Centered& [3,5]\\
Long&200&Centered& [21,22]\\
\bottomrule
\end{tabular}
}
\end{table}

As visible in \myblue{Figure~\ref{Distribution-Time}}, the share of the computational time of HGS-CVRP dedicated to the exploration of the \textsc{Swap*} neighborhood does not exceed 32\%, confirming the effectiveness of the proposed exploration and move filtering strategies which permit to limit the complexity of this neighborhood search. This is a significant achievement, given that this neighborhood is much larger than classic neighborhoods (e.g., \textsc{Relocate} or \textsc{Swap}).
According to our experiments, \textsc{Swap*} is responsible for approximately 15\% of the solution improvements. It also produces a larger proportion of improvements at later stages of the local search, when it is more difficult to find improving moves.

Some instance characteristics directly impact the performance of the neighborhoods. Instances with a larger number of customers or a depot location in the corner have proportionally more \textsc{Swap*} moves due to the proposed move filtering strategy. Consequently, the CPU time consumption and the number of applied moves tend to be higher in these situations. The \textsc{Swap*} neighborhood is also more effective when the routes contain fewer customers on average, as it finds a larger share of improvements in a smaller computational effort in proportion. Indeed, improving \textsc{Swap*} moves often correspond to pairs of \textsc{Relocate} moves that are individually improving but infeasible due to capacity limits, a recurrent case when optimizing many short routes.

\section{Conclusions}
\label{section-conclusions}

This paper helped to fulfill two important goals: 
1) facilitating future research by providing a simple, state-of-the-art code base for the CVRP; and
2) introducing a neighborhood called \textsc{Swap*} along with pruning and exploration strategies that permit an efficient search in a time similar to smaller neighborhoods such as \textsc{Swap}, \textsc{Relocate} or \textsc{2-Opt*}. We conducted an extensive computational campaign to compare the convergence of the proposed HGS-CVRP algorithm with that of the original HGS \citep{Vidal2012} and other algorithms provided to us by their authors, using the same computational platform. These analyses demonstrate that HGS-CVRP stands as the leading metaheuristic in terms of solution quality and convergence speed while remaining conceptually simple. As visible in these results, \textsc{Swap*} largely contribute to the search performance, especially at later stages of the local searches when improving moves are more difficult to find.

For future research, we recommend pursuing general research on the complexity of neighborhood explorations. For the CVRP, we can indeed argue that recent improvements in heuristic solution methods are due to more efficient and focused local searches rather than new metaheuristic concepts. Moreover, new breakthroughs are still being regularly made on the theoretical aspects of neighborhood search \citep[see, e.g.,][]{DeBerg2016a}. We also insist on the role of simplification in experimental design and scientific reasoning. While revisiting HGS and specializing it to the CVRP, we did systematic tests to retain only the most essential elements. Indeed, solution quality gains can almost always be achieved at the cost of having a method that is more convoluted and difficult to reproduce. Certainly, HGS-CVRP could be improved by dedicating a small fraction of its search effort into an additional set-partitioning component, additional decomposition phases, or mutation steps inspired by ruin-and-recreate. Nevertheless, we refrained from pushing further in that direction yet, as the goal of heuristic design should be to 1) identify methodological concepts that are as simple and effective as possible, and 2) to properly understand the role of each component. Therefore, we hope that the release of HGS-CVRP will contribute towards a general better understanding of heuristics for difficult combinatorial optimization problems.

\section*{Acknowledgments}

The author would like to acknowledge \myblue{Lucas Accorsi,} Florian Arnold, \myblue{Keld Helsgaun,} Fernando Obed, and Anand Subramanian for kindly giving access to their code for the experiments of this paper. Additional thanks are due to Alberto Santini for many useful C++ tips \myblue{as well as to two referees for their insightful recommendations.} This research has been partially funded by CAPES, CNPq [grant number 308528/2018-2], and FAPERJ [grant number E-26/202.790/2019] in Brazil. The computational resources have been provided by Calcul Québec and Compute Canada. This support is gratefully acknowledged.


\begin{thebibliography}{27}
\expandafter\ifx\csname natexlab\endcsname\relax\def\natexlab#1{#1}\fi
\expandafter\ifx\csname url\endcsname\relax
  \def\url#1{{\tt #1}}\fi
\expandafter\ifx\csname urlprefix\endcsname\relax\def\urlprefix{URL }\fi
\expandafter\ifx\csname urlstyle\endcsname\relax
  \expandafter\ifx\csname doi\endcsname\relax
  \def\doi#1{doi:\discretionary{}{}{}#1}\fi \else
  \expandafter\ifx\csname doi\endcsname\relax
  \def\doi{doi:\discretionary{}{}{}\begingroup \urlstyle{rm}\Url}\fi \fi

\bibitem[{Accorsi and Vigo(2021)}]{Accorsi2021}
Accorsi, L., D. Vigo. 2021.
\newblock {A fast and scalable heuristic for the solution of large-scale capacitated vehicle routing problems}.
\newblock {\it Transportation Science\/} {\bf 55}(4) 832--856.

\bibitem[{Aiex et~al.(2007)Aiex, Resende, and Ribeiro}]{Aiex2007}
Aiex, R.M., M.G.C. Resende, C.C. Ribeiro. 2007.
\newblock {TTT plots: A Perl program to create time-to-target plots}.
\newblock {\it Optimization Letters\/} {\bf 1}(4) 355--366.

\bibitem[{Arnold and S{\"{o}}rensen(2018)}]{Arnold2018d}
Arnold, F., K.~S{\"{o}}rensen. 2018.
\newblock {What makes a VRP solution good? The generation of problem-specific
  knowledge for heuristics}.
\newblock {\it Computers {\&} Operations Research\/} {\bf 106} 280--288.

\bibitem[{Beasley(1983)}]{Beasley1983}
Beasley, J.E. 1983.
\newblock {Route first-cluster second methods for vehicle routing}.
\newblock {\it Omega\/} {\bf 11}(4) 403--408.

\bibitem[{Christiaens and {Vanden Berghe}(2020)}]{Christiaens2019}
Christiaens, J., G.~{Vanden Berghe}. 2020.
\newblock {Slack induction by string removals for vehicle routing problems}.
\newblock {\it Transportation Science\/} {\bf 54}(2) 299--564.

\bibitem[{{De Berg} et~al.(2021){De Berg}, Buchin, Jansen, and
  Woeginger}]{DeBerg2016a}
{De Berg}, M., K.~Buchin, B.M.P. Jansen, G.~Woeginger. 2021.
\newblock {Fine-grained complexity analysis of two classic TSP variants}.
\newblock {\it ACM Transactions on Algorithms\/}. {\bf 17}(1) 1--29.

\bibitem[{Glover and Hao(2011)}]{Glover2011}
Glover, F., J.-K. Hao. 2011.
\newblock {The case for strategic oscillation}.
\newblock {\it Annals of Operations Research\/} {\bf 183}(1) 163--173.

\bibitem[{Guillen et~al.(2020)Guillen, Gendreau, Potvin, and Vidal}]{Reyes2020}
Guillen, F.O., M.~Gendreau, J.-Y. Potvin, T.~Vidal. 2020.
\newblock {A ruin {\&} recreate ant colony system algorithm for the capacitated
  vehicle routing problem. Working Paper. CIRRELT}.

\bibitem[{Helsgaun(2017)}]{Helsgaun2017}
Helsgaun, K. 2017.
\newblock {An extension of the Lin-Kernighan-Helsgaun TSP solver for constrained traveling salesman and vehicle routing problems}.
\newblock {Technical Report, Roskilde University}.

\bibitem[{Kendall et~al.(2016)Kendall, Bai, B{\l}azewicz, {De Causmaecker},
  Gendreau, John, Li, McCollum, Pesch, Qu, Sabar, {Vanden Berghe}, and
  Yee}]{Kendall2016a}
Kendall, G., R.~Bai, J.~B{\l}azewicz, P.~{De Causmaecker}, M.~Gendreau,
  R.~John, J.~Li, B.~McCollum, E.~Pesch, R.~Qu, N.~Sabar, G.~{Vanden Berghe},
  A.~Yee. 2016.
\newblock {Good laboratory practice for optimization research}.
\newblock {\it Journal of the Operational Research Society\/} {\bf 67}
  676--689.

\bibitem[{Laporte et~al.(2014)Laporte, Ropke, and Vidal}]{Laporte2014a}
Laporte, G., S.~Ropke, T.~Vidal. 2014.
\newblock {Heuristics for the vehicle routing problem}.
\newblock P.~Toth, D.~Vigo, eds., {\it Vehicle Routing: Problems, Methods, and
  Applications\/}, chap.~4. Society for Industrial and Applied Mathematics,
  87--116.

\bibitem[{Moscato and Cotta(2010)}]{Moscato2010}
Moscato, P., C.~Cotta. 2010.
\newblock {A modern introduction to memetic algorithms}.
\newblock M.~Gendreau, J.-Y. Potvin, eds., {\it Handbook of Metaheuristics\/},
  vol. 146. Springer, New York, 141--183.

\bibitem[{Oliver et~al.(1987)Oliver, Smith, and Holland}]{Oliver1987}
Oliver, I., D.~Smith, J.R. Holland. 1987.
\newblock {A study of permutation crossover operators on the traveling salesman
  problem.}
\newblock J.~Grefenstette, ed., {\it Genetic Algorithms and their Applications:
  Proceedings of the Second International Conference\/}. 224--230.

\bibitem[{Prins(2004)}]{Prins2004}
Prins, C. 2004.
\newblock {A simple and effective evolutionary algorithm for the vehicle
  routing problem}.
\newblock {\it Computers {\&} Operations Research\/} {\bf 31}(12) 1985--2002.

\bibitem[{Subramanian et~al.(2013)Subramanian, Uchoa, and
  Ochi}]{Subramanian2013}
Subramanian, A., E.~Uchoa, L.S. Ochi. 2013.
\newblock {A hybrid algorithm for a class of vehicle routing problems}.
\newblock {\it Computers {\&} Operations Research\/} {\bf 40}(10) 2519--2531.

\bibitem[{Talbi(2009)}]{Talbi2009}
Talbi, El~Ghazali. 2009.
\newblock {\it {Metaheuristics: From Design to Implementation}\/}.
\newblock John Wiley {\&} Sons.

\bibitem[{Toth and Vigo(2014)}]{Toth2014}
Toth, P., D.~Vigo, eds. 2014.
\newblock {\it {Vehicle Routing: Problems, Methods, and Applications}\/}.
\newblock 2nd ed. Society for Industrial and Applied Mathematics, Philadelphia.

\bibitem[{Uchoa et~al.(2017)Uchoa, Pecin, Pessoa, Poggi, Subramanian, and
  Vidal}]{Uchoa2017}
Uchoa, E., D.~Pecin, A.~Pessoa, M.~Poggi, A.~Subramanian, T.~Vidal. 2017.
\newblock {New benchmark instances for the capacitated vehicle routing
  problem}.
\newblock {\it European Journal of Operational Research\/} {\bf 257}(3)
  845--858.

\bibitem[{Vidal(2016)}]{Vidal2016}
Vidal, T. 2016.
\newblock {Technical note: Split algorithm in O(n) for the capacitated vehicle
  routing problem}.
\newblock {\it Computers {\&} Operations Research\/} {\bf 69} 40--47.

\bibitem[{Vidal(2017)}]{Vidal2017b}
Vidal, T. 2017.
\newblock {Node, edge, arc routing and turn penalties: Multiple problems -- One
  neighborhood extension}.
\newblock {\it Operations Research\/} {\bf 65}(4) 992--1010.

\bibitem[{Vidal et~al.(2012)Vidal, Crainic, Gendreau, Lahrichi, and
  Rei}]{Vidal2012}
Vidal, T., T.G. Crainic, M.~Gendreau, N.~Lahrichi, W.~Rei. 2012.
\newblock {A hybrid genetic algorithm for multidepot and periodic vehicle
  routing problems}.
\newblock {\it Operations Research\/} {\bf 60}(3) 611--624.

\bibitem[{Vidal et~al.(2014)Vidal, Crainic, Gendreau, and Prins}]{Vidal2012b}
Vidal, T., T.G. Crainic, M.~Gendreau, C.~Prins. 2014.
\newblock {A unified solution framework for multi-attribute vehicle routing
  problems}.
\newblock {\it European Journal of Operational Research\/} {\bf 234}(3)
  658--673.

\bibitem[{Vidal et~al.(2015)Vidal, Crainic, Gendreau, and Prins}]{Vidal2013a}
Vidal, T., T.G. Crainic, M.~Gendreau, C.~Prins. 2015.
\newblock {Time-window relaxations in vehicle routing heuristics}.
\newblock {\it Journal of Heuristics\/} {\bf 21}(3) 329--358.

\bibitem[{Vidal et~al.(2020)Vidal, Laporte, and Matl}]{Vidal2020}
Vidal, T., G.~Laporte, P.~Matl. 2020.
\newblock {A concise guide to existing and emerging vehicle routing problem
  variants}.
\newblock {\it European Journal of Operational Research\/} {\bf 286} 401--416.

\bibitem[{Vidal et~al.(2016)Vidal, Maculan, Ochi, and Penna}]{Vidal2014}
Vidal, T., N.~Maculan, L.S. Ochi, P.H.V. Penna. 2016.
\newblock {Large neighborhoods with implicit customer selection for vehicle
  routing problems with profits}.
\newblock {\it Transportation Science\/} {\bf 50}(2) 720--734.

\bibitem[{Vidal et~al.(2021)Vidal, Martinelli, Pham, and H{\`{a}}}]{Vidal2021}
Vidal, T., R.~Martinelli, T.A. Pham, M.H. H{\`{a}}. 2021.
\newblock {Arc routing with time-dependent travel times and paths}.
\newblock {\it Transportation Science\/} {\bf 55}(3) 706--724.

\bibitem[{Zachariadis and Kiranoudis(2010)}]{Zachariadis2010b}
Zachariadis, E.E., C.T. Kiranoudis. 2010.
\newblock {A strategy for reducing the computational complexity of local
  search-based methods for the vehicle routing problem}.
\newblock {\it Computers {\&} Operations Research\/} {\bf 37}(12) 2089--2105.

\end{thebibliography}

\end{document}